\documentclass[pdflatex,iicol,sn-basic]{sn-jnl}
\usepackage{algorithm}
\usepackage{algorithmic}
\usepackage{graphicx}
\usepackage{multirow}
\usepackage{amsmath,amssymb,amsfonts}
\usepackage{booktabs}
\usepackage{xurl}
\usepackage{array}
\usepackage{pdfpages}
\newcommand{\rqconclusion}[1]{%
  \par\smallskip
  \noindent\textbf{#1}\par
  \smallskip
}

\begin{document}

\title[Perspective-Bounded Evidence Calibration for Multimodal ToM]{%
Who Knows What---and What Does It Support? Perspective-Bounded Evidence
Calibration for Multimodal Theory of Mind}

\author[1]{\fnm{Ziqi} \sur{Liao}}
\author[1]{\fnm{Jiaqi} \sur{Shen}}
\author[1]{\fnm{Yiming} \sur{Na}}
\author[1]{\fnm{Kedi} \sur{Lyu}}
\author[2]{\fnm{Shao-Yuan} \sur{Lo}}
\author*[1]{\fnm{Hongxia} \sur{Xie}}
\email{hongxiaxie@jlu.edu.cn}

\affil[1]{\orgname{Jilin University},
\orgaddress{\city{Changchun}, \state{Jilin}, \country{China}}}

\affil[2]{\orgname{National Taiwan University},
\orgaddress{\city{Taipei}, \country{Taiwan}}}

\abstract{
A fundamental challenge in understanding social interactions is
determining how observable events support inferences about
individuals' latent mental states. In multimodal Theory of Mind
(ToM), the evidential value of an event depends on who has access
to it, how the information is acquired, and which mental state
is being inferred.
We formulate this challenge through \emph{evidence admissibility},
establishing an explicit correspondence between character-specific
information accessibility, source provenance, and claim-dependent
evidential relevance.
We introduce PIP-ToM, a perspective-bounded evidence calibration
framework that constructs agent-specific evidence stores from
reconstructed event traces, represents candidate mental states as
structured claims, and applies claim-aware adjustments to selected
local scores while preserving the underlying latent-variable
inference structure.
With GPT-4o, PIP-ToM achieves accuracies of 92.56\%, 95.00\%,
and 95.00\% on MuMA-ToM, MMToM-QA, and the filtered length-1
subset of Hi-ToM, respectively.
Controlled event-trace counterfactual evaluations show that
perspective storage improves belief accuracy under unobserved
relocations from 22\% to 99\%, with both configurations achieving
100\% world-state accuracy.
Further analyses reveal the complementary roles of structured
evidence evaluation and model-based inference in reasoning about
beliefs, goals, and recursively attributed mental states.}


\maketitle

\section{Introduction}
\label{sec:intro}

\begin{figure*}[t]
    \centering
    \includegraphics[width=0.9\textwidth]{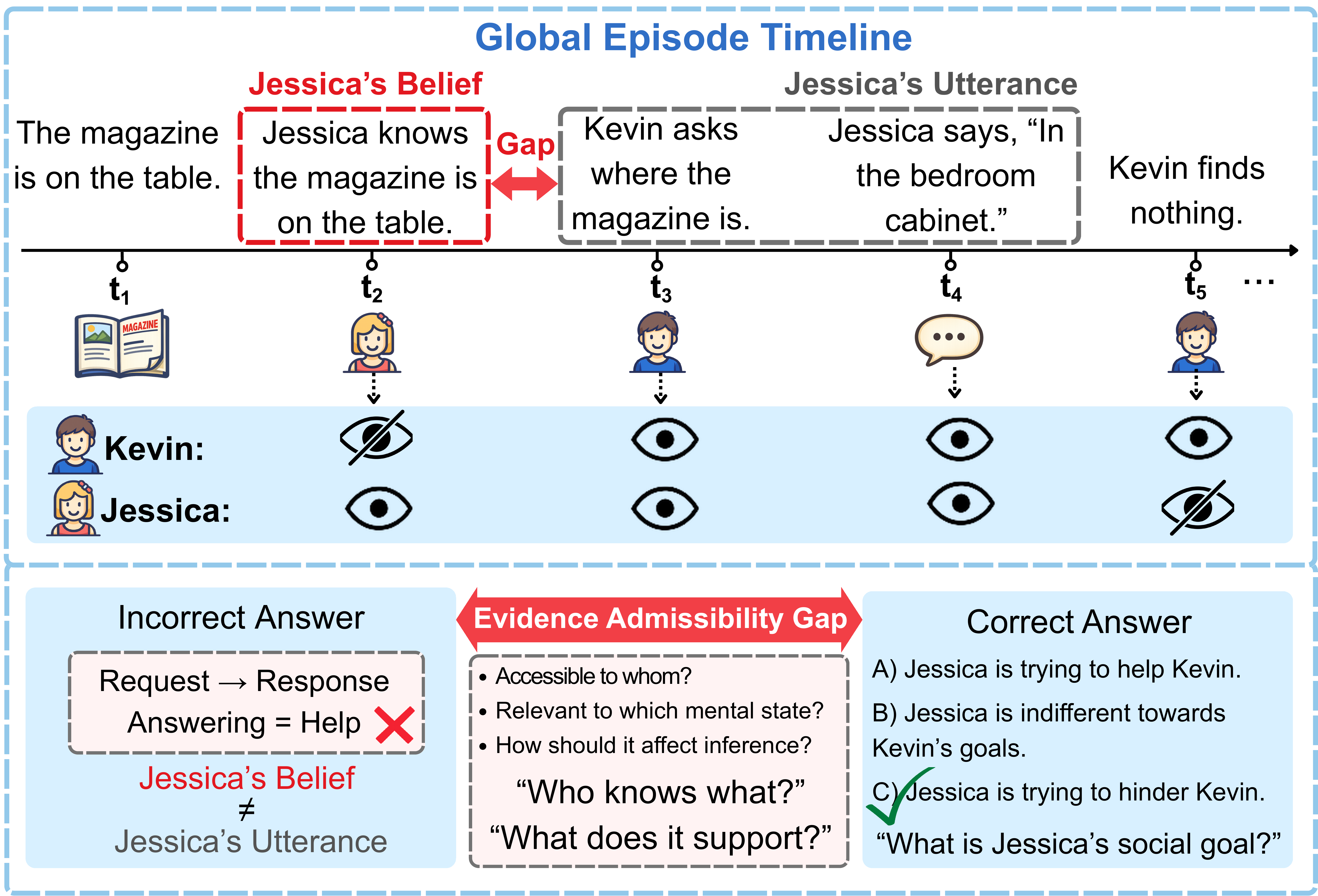}
    \caption{
The evidential roles of an utterance in mental-state inference.
The speaker's belief provides essential context for interpreting
her social intention, while the same utterance serves as a
source-attributed report for the listener.
}
    \label{fig:intro}
\end{figure*}

Theory of Mind (ToM), the ability to infer others' mental states,
including their beliefs, goals, and intentions, is fundamental to
understanding human behavior and social interaction
~\citep{premack1978does,Baker2017RationalAttribution}.
Multimodal ToM extends this capability to computational systems
that need to interpret visual events, human actions, and dialogue
in relation to the underlying mental states of the people involved.
Such inferences are important for social video understanding,
intention prediction, and human--robot collaboration, where an
intelligent system must anticipate human behavior, recognize mistaken
beliefs, and infer implicit goals
~\citep{dissing2020implementing,Zhang2025MindPowerET}.
Recent multimodal ToM benchmarks have formalized these challenges
through tasks involving physical beliefs, object-directed goals,
social intentions, and recursively attributed mental states
~\citep{jin2024mmtom,Shi2024MuMA_ToM}.
A fundamental difficulty across these settings is that individuals
may acquire different information from the same interaction,
giving rise to distinct beliefs and interpretations of observed
events~\citep{Kim2023FANToMAB,WimmerPerner1983Beliefs}.

A complete observation of an interaction may reveal events that
individual participants never perceive, while the same accessible
event may support different conclusions about their mental states.
Figure~\ref{fig:intro} illustrates this challenge through a social
interaction in which Jessica tells Kevin to look for a magazine
in the bedroom cabinet.
Kevin follows her advice and finds nothing.
If Jessica sincerely believes that the magazine is in the cabinet,
her advice is consistent with a mistaken attempt to help.
However, given her belief that the magazine is on the table,
the same utterance provides stronger evidence of deliberate
misdirection.
Furthermore, the utterance has different evidential roles for
the two participants: it constitutes a communicative action that
may reveal Jessica's social intention and a source-attributed
report that may inform Kevin's belief and subsequent search
~\citep{GoodmanFrank2016Pragmatic}.
Understanding such interactions therefore requires addressing
two closely related questions:
\emph{Who had access to the evidence, and what mental-state
claim can that evidence support?}

Existing approaches have made substantial progress in reconstructing
agent perspectives and inferring latent mental states.
Perspective-taking methods such as SimToM filter episode information
according to the target character's perspective
~\citep{Wilf2023ThinkTP}, while PercepToM explicitly distinguishes
perception inference from subsequent belief inference
~\citep{jung-etal-2024-perceptions}.
Bayesian inverse-planning approaches evaluate competing mental-state
hypotheses by examining how well they explain observed actions and
social behavior~\citep{jin2024mmtom,Shi2024MuMA_ToM}.
AutoToM further improves the flexibility of model-based inference
by automatically constructing task-relevant latent variables
and their dependencies~\citep{zhang2025autotom}.
Together, these advances motivate a complementary question at
the interface between evidence reconstruction and mental-state
inference:
\textit{how should a reconstructed event be associated with
a particular character, mental-state claim, and local inference
score?}
Explicitly modeling this correspondence is particularly important
when objective states diverge from character beliefs or when
observed actions and utterances admit multiple mental-state
explanations.

We formulate this challenge in terms of
\textbf{evidence admissibility}, which characterizes the conditions
under which reconstructed evidence can legitimately contribute
to the evaluation of a mental-state claim.
Three complementary properties govern this correspondence.
\emph{Accessibility} determines whether a record is available
to the relevant character.
\emph{Provenance} specifies how the information was acquired,
such as through direct observation, personal action, or another
person's report.
\emph{Claim-dependent relevance} determines which mental-state
hypothesis the record can support or contradict.
For instance, observing an empty cabinet provides negative
evidence about an object's location, whereas a cabinet that
a character has never inspected remains an unresolved possibility.
Similarly, hearing an utterance establishes access to the
communicated information, while its implications for the
speaker's intention depend on the relationship between the
utterance, the speaker's belief, and the anticipated social effect.
These distinctions provide an evidential basis for reasoning
about beliefs, goals, and intentions under asymmetric information
access.

To operationalize this formulation, we introduce
\textbf{PIP-ToM}, a perspective-bounded evidence calibration
framework for multimodal Theory of Mind.
Given a temporally ordered event trace reconstructed from
the available multimodal inputs, PIP-ToM first constructs
\emph{agent-specific perspective stores} that preserve accessible
observations, actions, utterances, and their information sources.
It then represents candidate answers as \emph{structured mental-state
claims}, making explicit the queried mental-state type, relevant
characters, entities, and candidate values.
Based on these representations, \emph{claim-aware evidence routing}
associates accessible records with the corresponding perceptual,
behavioral, communicative, and recursive evidence relations.
The resulting claim--evidence compatibility is mapped to bounded
factors that reweight selected local scores within an underlying
model-based inference procedure.
Through this design, PIP-ToM explicitly connects character-relative
evidence with candidate mental-state evaluation while preserving
the latent-variable structure and marginalization procedure of
the base model.

We evaluate PIP-ToM on MuMA-ToM, MMToM-QA, and the filtered
length-1 subset of Hi-ToM following AutoToM's candidate-filtering
protocol~\citep{He2023HiTOM,zhang2025autotom}.
With GPT-4o, PIP-ToM achieves accuracies of 92.56\%, 95.00\%,
and 95.00\%, respectively.
To directly examine perspective-sensitive evidence use, we
construct a controlled event-trace counterfactual evaluation
based on 50 Hi-ToM scenarios, independently manipulating
object relocation destinations and character-specific access.
Adding perspective storage improves belief accuracy under
unobserved relocations from 22\% to 99\% and quadruplet accuracy
from 12\% to 96\%, while both configurations achieve 100\%
world-state accuracy.
Component analyses further reveal the importance of aligning
evidence with the queried mental-state claim.
Specifically, removing the structured claim representation
and its associated reweighting reduces MuMA-ToM Social Goal
accuracy from 84.33\% to 67.67\%.
Finally, on the same 300 MuMA-ToM Belief-of-Goal questions
covered by direct evidence matching, full PIP-ToM improves
accuracy from 76.00\% to 96.67\%.
Together, these findings characterize the complementary roles
of information accessibility, claim-aware evidence use, and
model-based inference in different mental-state reasoning tasks.




Our main contributions are summarized as follows:

\begin{itemize}
    \item We formulate \textbf{evidence admissibility} for multimodal Theory of Mind, characterizing the relationship between character-specific information accessibility,
    source provenance, and claim-dependent evidential relevance.

    \item We introduce \textbf{PIP-ToM}, a modular framework that integrates source-preserving perspective storage  and structured mental-state claims with claim-aware score adjustments in model-based inference.

    \item We develop a \textbf{controlled counterfactual
    evaluation} that independently manipulates objective
    world-state changes and character-specific access.
    Together with benchmark comparisons, component ablations,
    and direct-matching diagnostics, it provides empirical
    insights into evidence use across belief, goal, and
    recursive mental-state reasoning tasks.

\end{itemize}
\section{Related Work}

\subsection{Explicit State Tracking}

Explicit state reconstruction gives ToM solvers a structured account of how
events change the environment~\citep{Sclar2024ExploreToM}. Discrete World Models~\citep{Huang2024ANO}
describe state transitions induced by agents' interactions and relate task
complexity to the states needed for a solution. Such representations reduce
the burden of implicitly reconstructing a changing world from an unstructured
narrative. EnigmaToM~\citep{Xu2025EnigmaToMIL} combines a neural knowledge base
of entity states with spatial scene graphs, iterative perspective masking,
and knowledge injection. It therefore connects entity-state reconstruction
with belief tracking and perspective taking, rather than treating world
modeling as a purely agent-independent preprocessing step.

Symbolic approaches further make state updates explicit and checkable~\citep{dissing2020implementing}.
PDDL-Mind~\citep{Zhu2026PDDLMindLL} translates narratives into states and
actions expressed in the Planning Domain Definition Language and verifies
action-induced transitions against a predefined domain. Recording access
through predicates such as \emph{seen} and \emph{heard} also provides a basis
for distinguishing events in the world from events available to an agent.


These approaches transform unstructured narratives into
explicit world-state or belief-state trajectories. By externalizing entities,
actions, temporal transitions, and information-access markers, they reduce
the need for a solver to repeatedly reconstruct the evolving environment
from raw narrative context. The resulting representations also make
intermediate state updates more inspectable and provide a structured basis
for subsequent belief and knowledge inference. Nevertheless, state tracking
primarily establishes what occurred and how the environment changed; further
reasoning is required to determine how reconstructed events relate to the
information available to each individual agent.



\subsection{Perspective Reconstruction}

Perspective reconstruction explicitly separates the solver's information from the information available to a character~\citep{Thiyagarajan2025UniToMBench}. SimToM~\citep{Wilf2023ThinkTP} first filters events according to the target character's perspective and then answers using the resulting context. PercepToM~\citep{jung-etal-2024-perceptions} distinguishes perception inference from perception-to-belief inference, making information access an explicit intermediate step and examining how perceived information is translated into beliefs. Both approaches recognize that the complete narrative available to the solver may contain facts that the target agent never observed, thereby establishing information accessibility as an important condition for mental-state inference~\citep{Kim2023FANToMAB,Fan2025SoMiToM,jeon-lee-2026-picture}.

Beyond reconstructing accessible contexts, other methods maintain perspectives as evolving representations. SymbolicToM~\citep{sclar-etal-2023-minding} uses a multi-character belief tracker with graph-based representations, while TimeToM~\citep{Hou2024TimeToM} constructs temporal belief-state chains and distinguishes beliefs about the world from beliefs about other agents. ThoughtTracing~\citep{Kim2025HypothesisDrivenTR} generates, propagates, and reweights natural-language hypotheses about an agent's evolving mental state using its perceptions and actions. These methods extend perspective taking from context filtering to the temporal tracking of observations, belief updates, and attributed mental states, which is particularly important under asymmetric information access, false beliefs, and nested attribution~\citep{Xiao2025DynToM,le-etal-2019-revisiting,xu-etal-2024-opentom,He2023HiTOM}.



However, reconstructing an agent-relative context primarily determines which evidence was accessible, rather than what each accessible event can support. The same action, observation, or utterance may have different evidential roles depending on whether the queried variable concerns a physical belief, an object-directed goal, a communicative effect, or a recursively attributed mental state~\citep{Gandhi2023UnderstandingSR,GoodmanFrank2016Pragmatic}.

\subsection{Bayesian and Model-Based Mental-State Inference}

Bayesian ToM formalizes mentalizing as inverse inference over the mechanisms
that generate behavior. Classical inverse-planning
models~\citep{Baker2009InversePlanning,Baker2017RationalAttribution,JaraEttinger2019IRL}
relate actions to beliefs, desires, and percepts, allowing observed behavior
to constrain several latent mental variables jointly. This formulation is
particularly relevant when an agent's action is rational under its belief even
though that belief differs from the objective world state. In multimodal
settings, BIP-ALM~\citep{jin2024mmtom} augments Bayesian inverse planning with
language-model likelihood estimates, while LIMP~\citep{Shi2024MuMA_ToM}
uses language-based inverse planning to reason about beliefs, goals, and social
interactions. Work on scalable Bayesian
planning~\citep{zhang2025overcomingmultistepcomplexitymultimodal} further
addresses the computational demands of multi-step multimodal inference.

Recent methods broaden how hypotheses and model structure are obtained.
AutoToM~\citep{zhang2025autotom} automates the selection of task-relevant
mental variables and the construction of their dependencies, reducing the need
to manually specify a separate inference structure for each question family.
MindZero~\citep{zhang2026mindzerolearningonlinemental}
takes a learning-based direction, using planner-estimated action likelihoods
as self-supervised reinforcement-learning signals and internalizing mental
reasoning into single-pass inference without explicit mental-state annotations. 



Bayesian and model-based approaches treat beliefs, goals,
intentions, and percepts as latent variables that explain observed behavior.
They therefore move beyond direct answer generation by explicitly modeling
dependencies among mental states, actions, and observations~\citep{westby2023collectiveintelligencehumanaiteams,ying-etal-2025-understanding}. These approaches provide a
principled account of how behavior constrains competing mental-state
hypotheses once the relevant variables, dependencies, and local evidence
scores have been specified.

Across the three strands of work, explicit state tracking establishes what
occurred and how the environment evolved, perspective reconstruction
determines which parts of that history were available to each agent, and
model-based inference evaluates how latent mental states explain observed
behavior. A remaining interface, however, is not typically made explicit:
\textit{which agent-accessible evidence may legitimately affect which local
term of a candidate mental-state hypothesis?} Accessibility alone does not
determine evidential relevance, while probabilistic hypothesis evaluation
depends on how observations, actions, and utterances are assigned to its
local scores. PIP-ToM targets this interface by combining agent-specific
perspective storage with typed mental-state schemas and claim-dependent
evidence routes. It thereby constrains how reconstructed evidence enters
candidate evaluation while preserving the latent variables, dependencies,
and marginalization procedure of the underlying model.

\section{Methodology}

\begin{figure*}[t]
\centering
\includegraphics[width=\textwidth]{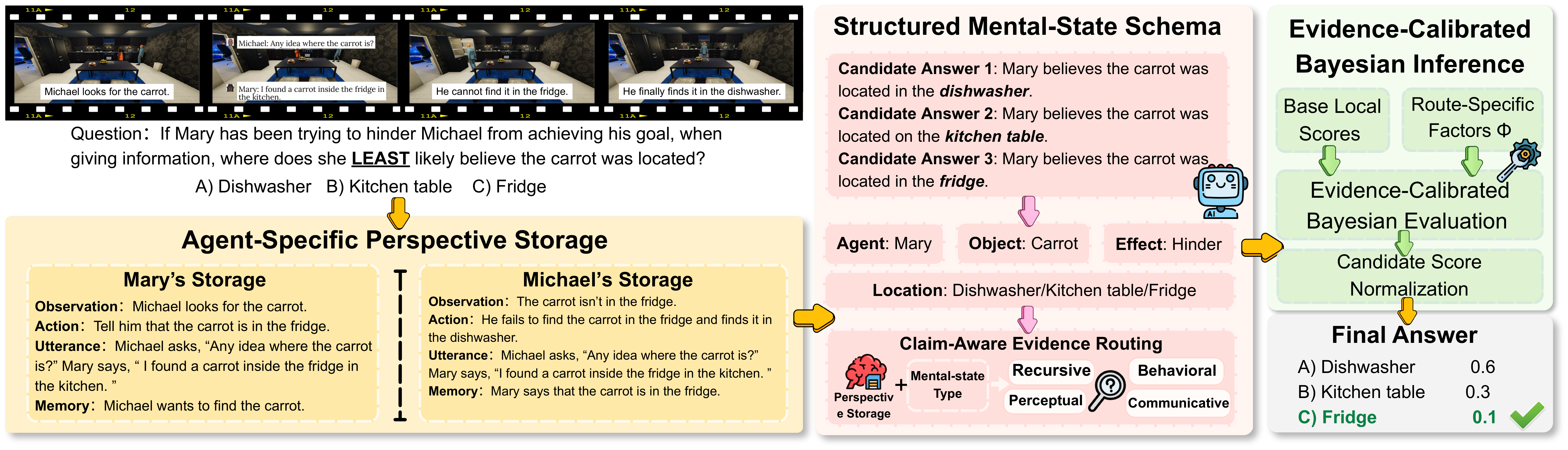}
\caption{Overview of PIP-ToM. It transforms a fused episode trace and candidate answers into perspective-bounded evidence, route-specific factors, and normalized candidate scores for answer selection.}
\label{fig:framework}
\end{figure*}

\subsection{Evidence Admissibility Formulation}

We consider multimodal Theory of Mind (ToM) inference from a temporally
ordered fused episode trace $E_{1:T}$. A latent-variable base model
constructs mental-state hypotheses and provides local surrogate scores
$L_{k,n}^{\mathrm{base}}$ for candidate $k$ and inference term $n$.
However, a globally reconstructed trace may contain events that the
relevant perspective holder could not access, events unrelated to the
mental-state variable being evaluated, or evidence whose source is
inappropriate for the claim. Allowing such evidence to influence a local
score may produce an answer supported by the wrong evidential basis.

We formulate this problem through an \textbf{Evidence Admissibility
Principle}: an event may contribute to a candidate-specific inference
term only if it is accessible to the relevant perspective holder, has
provenance compatible with the term, and is diagnostic of its
mental-state claim. For event $e_t$, perspective holder $i$, candidate
schema $Z_k$, and local inference term $n$, we write
\begin{equation}
\begin{aligned}
\operatorname{Adm}(e_t;i,Z_k,n)
={}& \operatorname{Acc}_i(e_t) \\
&\wedge \operatorname{Prov}_n(e_t,Z_k)
\wedge \operatorname{Rel}_n(e_t,Z_k).
\end{aligned}
\label{eq:evidence_admissibility}
\end{equation}
where $\operatorname{Acc}_i(e_t)$ indicates whether the relevant
perspective holder could access the event through perception,
communication, or its own action; $\operatorname{Prov}_n(e_t,Z_k)$
indicates whether the event's source is compatible with its use in term
$n$; and $\operatorname{Rel}_n(e_t,Z_k)$ indicates whether the event
bears on the candidate-specific mental-state variable evaluated by that
term. Provenance distinguishes, for example, utterances, observed
actions, and objective state changes: an utterance may provide evidence
of what was communicated without establishing that its content is true.
The relevant perspective and temporal scope depend on the inference
term. For example, solver-observed behavior may inform retrospective
goal inference without establishing that the agent had access to the
corresponding information at an earlier time.

Equation~\ref{eq:evidence_admissibility} separates evidence
\emph{availability}, \emph{source compatibility}, and
\emph{relevance}. Inaccessible events are excluded from the
agent-scoped evidence view. Accessible events with incompatible
provenance or insufficient relevance may remain in that view but must
not modify the corresponding local term. Evidence satisfying all three
conditions is eligible for further compatibility assessment;
admissibility alone does not imply support for a particular candidate.
PIP-ToM operationalizes this principle through agent-specific
perspective storage, source-preserving evidence representation,
structured mental-state claims, and claim-dependent calibration of the
base model's local scores.

\subsection{Overview of PIP-ToM}

Multimodal ToM requires reasoning about beliefs, goals, and intentions
from observations, actions, communication, and memory under partial
observability. Bayesian inverse planning
~\citep{Baker2009InversePlanning,Baker2017RationalAttribution,
JaraEttinger2019IRL} provides a model-based account of how latent mental
states can explain behavior. Its inference structure, however, does not by
itself specify whether each reconstructed event is an admissible source of
evidence for the queried agent and candidate claim.

Figure~\ref{fig:framework} illustrates PIP-ToM as an
\textbf{evidence-to-inference interface} built on an existing latent-state
model. First, Agent-Specific Perspective Storage maintains accessible,
source-preserving evidence for each agent. Second, a Structured
Mental-State Schema interprets the question and candidates as typed
claims, determining which evidence relations are relevant to the queried
mental state. Third, claim-aware routes evaluate those relations and
produce factors that locally calibrate the base model's scores. The
framework therefore makes explicit both \emph{who had access to the
evidence} and \emph{what that evidence may support}, without replacing the
latent variables or their dependencies.

Algorithm~\ref{alg:pip_end_to_end_main} summarizes the procedure. The
shared query frame and perspective stores are constructed once per
question. Each candidate then activates the applicable evidence routes,
whose outputs are composed at the relevant local terms. Marginalization
and answer selection follow the underlying model.

\begin{algorithm}[t]
\caption{End-to-end PIP-ToM inference.}
\label{alg:pip_end_to_end_main}
\begin{algorithmic}[1]
\REQUIRE Fused event trace $E_{1:T}$, question $q$, candidates $C_{1:K}$,
base latent-variable model $\mathcal{G}$
\ENSURE Normalized scores $\widetilde{P}_{1:K}$ and answer $\widehat{k}$
\STATE Extract agents and normalize $q$ and $C_{1:K}$
\STATE Parse the shared frame $R_q$ and candidate schemas $Z_{1:K}$
\STATE Construct agent-specific stores $\{\mathcal{P}_i^T\}$ from $E_{1:T}$
\FOR{candidate $k=1,\ldots,K$}
  \STATE Activate claim-dependent routes $\mathcal{M}_k$
  \FOR{local term $n$ in $\mathcal{G}$}
    \STATE Extract admissible, source-preserving evidence for active routes
    \STATE Evaluate claim--evidence compatibility and obtain $\Phi_{k,n}^{(m)}$
    \STATE Compute $\Phi_{k,n}=\prod_{m\in\mathcal{M}_{k,n}}
    \Phi_{k,n}^{(m)}$
    \STATE Set $L_{k,n}^{\mathrm{PIP}}
    =\min(1,L_{k,n}^{\mathrm{base}}\Phi_{k,n})$
  \ENDFOR
  \STATE Marginalize latent assignments of $\mathcal{G}$
\ENDFOR
\STATE Normalize candidate scores and decode the question operator $\rho$
\RETURN $\widetilde{P}_{1:K},\widehat{k}$
\end{algorithmic}
\end{algorithm}

\subsection{Agent-Specific Perspective Storage}

Agent-Specific Perspective Storage establishes the epistemic boundary
between the globally reconstructed episode and the evidence available to
a particular agent. In the Mary--Michael example in
Figure~\ref{fig:framework}, Mary's own utterance and request-related
observations belong to her perspective, whereas Michael's later failed
search and discovery belong to his perspective unless Mary could access
those events. A shared episode therefore gives rise to distinct,
agent-centered evidence views.

For every agent $i$ and timestep $t$, we maintain
\begin{equation}
\mathcal{P}_i^t
=
\{O_i^{1:t},A_i^{1:t},U_i^{1:t},M_i^t\},
\label{eq:perspective_storage}
\end{equation}
where $O_i^{1:t}$ contains accessible observations,
$A_i^{1:t}$ contains actions performed by or observed by $i$,
$U_i^{1:t}$ contains utterances produced or heard by $i$, and $M_i^t$
is source-aware memory formed from previously admitted evidence. Each
stored record retains available information about its actor or speaker,
temporal position, and source of access.

For an incoming event $e_t$, we express the access gate as
\begin{equation}
g_i^t =
\mathbf{1}\!\left[
v_i(e_t)\vee u_i(e_t)\vee a_i(e_t)
\right],
\label{eq:access_gate}
\end{equation}
where $v_i$, $u_i$, and $a_i$ indicate visibility,
audibility, and self-action cues, respectively. 
The store is updated according to
\begin{equation}
\mathcal{P}_i^t
=
\operatorname{Upd}\!\left(
\mathcal{P}_i^{t-1},g_i^t,e_t
\right).
\label{eq:perspective_update}
\end{equation}
The gate determines whether the event is admitted to an agent's evidence
view; it does not certify the truth of the event's semantic content. An
utterance is recorded as a communicative act of its speaker and, for an
audible listener, as a heard utterance. The asserted content is not
silently promoted to an objective world-state update. Likewise, observing
an action does not imply access to a later outcome that occurred outside
the agent's perspective.

This distinction also governs memory. Previously admitted observations,
actions, and utterances can persist across timesteps with their provenance
intact, but persistence does not create new perceptual access. Any
intermediate belief estimate may be retained as an inferred state, not
reintroduced as an observed event. The storage thus organizes the
\emph{available evidential history} rather than supplying oracle mental
states. Equations~\ref{eq:access_gate}--\ref{eq:perspective_update}
specify the interface-level semantics; the access cues are instantiated
from the supplied timeline fields and source tags. Detailed construction
and input normalization are given in Supplementary Sec.~A.1.

\subsection{Structured Mental-State Schema}

Perspective reconstruction determines \emph{which} events an agent could
access, but accessibility alone does not specify \emph{what} those events
can support. A search action can inform a goal hypothesis, a perceived
object location can inform a belief hypothesis, and an utterance can
inform a social-intention hypothesis only when interpreted in relation
to the speaker's belief and its anticipated effect. PIP-ToM therefore
represents each candidate as a typed mental-state claim and uses that
representation to determine how accessible evidence enters inference.
The schema and evidence routes form a single claim-aware interface:
the former identifies the queried relation, while the latter evaluates
its relevant evidence.

Given question $q$ and candidates $C_{1:K}$, the interface extracts a
shared query frame and a schema for each candidate:
\begin{equation}
(R_q,Z_{1:K})
=
\operatorname{Parse}\!\left(q,C_{1:K}\right),
\label{eq:schema_parse}
\end{equation}
where $\operatorname{Parse}$ combines fixed-prompt candidate normalization
with deterministic, task-aware extraction of structured fields. The
shared frame is held constant across candidates, avoiding a different
interpretation of the question for each answer option. The resulting
representations are
\begin{equation}
R_q=\langle\tau,i,j,x,\rho\rangle,
\qquad
Z_k=\langle R_q,\chi_k,\epsilon_k\rangle,
\label{eq:schema_roles}
\end{equation}
where $\tau$ denotes the mental-state type, $i$ the perspective holder,
$j$ the target agent when applicable, $x$ the queried entity, and $\rho$
the question relation or polarity. Here $\chi_k$ specifies the
candidate value and $\epsilon_k$ an optional behavioral or social
effect. Prompts, normalized fields, and task adapters are detailed in
Supplementary Sec.~A.1.

\paragraph{Claim-dependent evidence requirements.}
The question addressed by the schema is: \emph{what evidence can
legitimately support this particular mental-state claim?} The answer
depends on the claim's semantic structure, rather than on the presence
of a matching word or event in the complete episode. Belief
claims require evidence of states or changes accessible to the belief
holder; a fact from the global trace cannot establish what that agent
believes if the agent never encountered it. Goal claims additionally
require behavioral consistency: the same action may be compatible with
different goals depending on the actor's accessible information. Social-goal claims require a joint interpretation of communicative behavior,
the speaker's belief, and the intended effect on a listener; providing
a location is not by itself evidence of a helpful intention. Belief-of-goal and higher-order belief claims require nested attribution, in which
each step of reasoning respects whose information and behavior could be
known to the attributing agent. These requirements connect the
candidate's mental-state variable to the evidence relations that can
actually distinguish it from alternatives.

PIP-ToM implements these requirements through four complementary routes:
\begin{equation}
\mathcal{M}_k\subseteq
\left\{
\begin{aligned}
&\mathrm{Perceptual},\ \mathrm{Behavioral},\\
&\mathrm{Communicative},\ \mathrm{Recursive}
\end{aligned}
\right\},
\label{eq:route_set}
\end{equation}
with roles defined as follows:
\begin{itemize}
  \item \textbf{Perceptual evidence} relates accessible observations,
  state changes, and source-aware memories to the candidate's physical
  belief or other relevant state claim.
  \item \textbf{Behavioral evidence} relates observed or self-produced
  actions and trajectories to hypothesized goals and behavioral effects,
  interpreted under the agent's available information.
  \item \textbf{Communicative evidence} retains the speaker, listener,
  and content of utterances, and assesses their relation to the queried
  belief or intended communicative effect.
  \item \textbf{Recursive evidence} follows the attribution chain in
  nested claims, relating what one agent could know about another's
  observations, beliefs, or behavior to the candidate attribution.
\end{itemize}

\paragraph{Compositional routing and evidential roles.}
Routes are complementary evidence channels, not mutually exclusive
labels for question categories. A physical-belief claim is primarily
grounded in perceptual evidence, but an admissible action or utterance
may also be diagnostic. A goal claim combines behavior with the
perceptual context in which it occurred. A social-goal claim combines
communicative evidence with the speaker's perspective and possible
behavioral consequences. Recursive claims activate the attribution
route together with lower-level evidence needed to evaluate the inner
claim. The typed schema activates only routes and relations applicable
to the queried local term; unavailable optional evidence leaves that
term unchanged.

For local term $n$ and candidate $k$, let
$\mathcal{M}_{k,n}\subseteq\mathcal{M}_k$ denote the relevant routes.
Each route extracts admissible evidence from the agent's store and
returns a claim--evidence compatibility factor:
\begin{equation}
\begin{aligned}
E_k^{(m)}
&=\operatorname{Extract}_{m}\!\left(Z_k,\mathcal{P}_i^T\right),\\
\Phi_{k,n}^{(m)}
&=G_{m,n}\!\left(E_k^{(m)},Z_k,R_q\right),
\quad m\in\mathcal{M}_{k,n}.
\end{aligned}
\label{eq:evidence_factor}
\end{equation}
The source-preserving extractor excludes inaccessible events before
assessing compatibility. Its output distinguishes evidence that
\emph{supports} a candidate, \emph{contradicts} it, or leaves it
\emph{unresolved}. For example, a source-tagged utterance may support
a claim about what was communicated without establishing that the
uttered proposition is true in the world. An observed unsuccessful
search may contradict a location claim, whereas a location never
inspected by the agent is simply unknown to that agent. Thus, absent
accessible evidence is not automatically treated as negative evidence.

When multiple records are available, the implementation combines their
relations within the corresponding route. Alternative interpretations
of the same record are mutually exclusive, preventing one event from
being counted as both support and contradiction in a single local term.
Distinct, admissible sources can contribute complementary or conflicting
evidence without losing their provenance. The route-specific treatment
of these cases is detailed in Supplementary Secs.~A.2--A.3.

\subsection{Evidence-Calibrated Bayesian Inverse Planning}

The preceding components determine which evidence is admissible and how
it relates to each candidate. The final step is to make these structured
judgments effective within the underlying model's local hypothesis
evaluation, rather than replacing its mental-state inference with a
direct answer rule. We use \textbf{local evidence calibration}: for each
candidate $k$ and local term $n$, the outputs of applicable routes are
combined as
\begin{equation}
\Phi_{k,n}
=
\prod_{m\in\mathcal{M}_{k,n}}
\Phi_{k,n}^{(m)},
\label{eq:combined_factor}
\end{equation}
and applied to the base local surrogate score:
\begin{equation}
L_{k,n}^{\mathrm{PIP}}(\mathbf{h})
=
\min\!\left(
 c,\,
 L_{k,n}^{\mathrm{base}}(\mathbf{h})\Phi_{k,n}
\right).
\label{eq:calibrated_likelihood}
\end{equation}
Here $\mathbf{h}$ is a latent-variable assignment and
$L_{k,n}^{\mathrm{base}}(\mathbf{h})$ is the score supplied by the
underlying model. As in the backbone implementation, the local score is
obtained from a binary-response probability and lies in $[0,1]$; we
set $c=1$. If no route is applicable, the empty product gives
$\Phi_{k,n}=1$, and the local term is unchanged.

\paragraph{Calibration design.}
Three constraints govern how claim--evidence compatibility enters local
scoring. \textbf{First, admissibility precedes adjustment:} evidence
outside the relevant agent's perspective, or unrelated to the mental-state variable represented by the term, cannot change its calibration
factor. \textbf{Second, compatibility determines adjustment direction:}
explicit support receives a factor greater than $1$, contradiction
receives a positive factor below $1$, and missing, uncertain, or
non-diagnostic evidence receives the neutral factor $1$. Observed
negative evidence, such as an unsuccessful search, is distinguished
from an unobserved or unknown state. \textbf{Third, adjustment is local:}
a route changes only the local terms to which its evidence relation
applies. The rest of the latent-variable graph, including unrelated
terms, remains intact. Supporting factors can increase a local score
only up to the cap $c$, while downweighting factors reduce the score
without redefining the underlying hypothesis space.

Detailed factor configurations and composition rules are provided in Supplementary Secs.~A.2--A.3, while Sec.~\ref{sec:rq2} examines sensitivity to calibration strength.

\paragraph{Calibration in Model-Based Inference.}

By composing route-specific adjustments before latent-variable marginalization, PIP-ToM allows admissible evidence to influence candidate evaluation without altering the underlying inference structure. The multiplicative composition represents heuristic score adjustments rather than a product of independently estimated probabilities and does not imply additional conditional-independence assumptions. Following the base model's marginalization procedure, we obtain the normalized candidate scores:
\begin{equation}
\widetilde{P}_k =
\frac{\sum_h P(h)\prod_n L^{\mathrm{PIP}}_{k,n}(h)}
{\sum_{k'}\sum_h P(h)\prod_n L^{\mathrm{PIP}}_{k',n}(h)}.
\label{eq:piptom_posterior}
\end{equation}
Here $k$ indexes the candidate being evaluated, and $k'$
indexes all candidates. The resulting $\widetilde{P}_k$ is a normalized, evidence-calibrated candidate score and is not necessarily a statistically calibrated posterior probability. The question operator determines the final answer:
\begin{equation}
\widehat{k}
=
\begin{cases}
\operatorname*{arg\,max}_{k}\widetilde{P}_k,
& \rho=\mathrm{Most},\\
\operatorname*{arg\,min}_{k}\widetilde{P}_k,
& \rho=\mathrm{Least}.
\end{cases}
\label{eq:answer_decoding}
\end{equation}

PIP-ToM thus preserves the latent variables, dependencies, priors, marginalization, and answer selection procedure of the base inference model while constraining the evidential basis of its local scores. Inaccessible events are excluded from agent-scoped evidence views, accessible but irrelevant evidence remains neutral, and diagnostic evidence contributes through the corresponding claim-aware routes. Rather than imposing hard invariance on the final predictions, PIP-ToM provides an explicit evidence-admissibility mechanism whose effectiveness is evaluated through controlled counterfactual experiments in Sec.~\ref{sec:counterfactual}.

\section{Experiments}

\subsection{Benchmarks}

We evaluate PIP-ToM on three complementary ToM benchmarks:
MuMA-ToM~\citep{Shi2024MuMA_ToM}, MMToM-QA~\citep{jin2024mmtom},
and Hi-ToM~\citep{He2023HiTOM}. 

\paragraph{MuMA-ToM}
MuMA-ToM~\citep{Shi2024MuMA_ToM} is a multimodal benchmark grounded in the VirtualHome simulator~\citep{puig2018virtualhomesimulatinghouseholdactivities}.
It contains 225 household social-interaction episodes from which 900 multiple-choice questions are constructed, with 300 questions each for physical belief, social goal, and belief-of-goal inference. These tasks
require the model to integrate perception, action, and communication while distinguishing the objective episode state from the information available to each agent.

\paragraph{MMToM-QA}
MMToM-QA~\citep{jin2024mmtom} is also grounded in VirtualHome and combines video with textual descriptions, but centers on household object-search episodes. It is built from 134 videos and contains 600 two-choice questions,
evenly divided into 300 belief questions and 300 goal questions. These questions cover true, false, and updated beliefs as well as evidence from future actions. The benchmark therefore tests whether a model can track how
an agent's observations and search history constrain both its current belief and the actions consistent with its latent goal.

\paragraph{Hi-ToM}
Hi-ToM~\citep{He2023HiTOM} complements the two multimodal benchmarks with text-based multi-agent stories involving object movements, agent observations, and communication. Following the evaluation protocol used by AutoToM, we evaluate the length-1 subset, which contains 200 questions distributed evenly
across reasoning orders 0--4, with 40 questions at each order. These questions
range from objective world states to deeply nested beliefs about other agents' beliefs and require models to track agent-specific information across events and correctly propagate beliefs through multiple levels of recursive mental-state reasoning.




Detailed descriptions of the task settings, dataset sizes, and model configurations are provided in Supplementary Sec.~B.



\begin{table*}[t]
\centering
\small
\setlength{\tabcolsep}{1.5pt}
\renewcommand{\arraystretch}{0.93}
\caption{
\textbf{Accuracy (\%) on MuMA-ToM, MMToM-QA, and Hi-ToM.} BoG denotes Belief of Goal; ``--'' indicates that the result is unavailable, not reported, or not applicable to the benchmark. \textbf{Bold} and \underline{underlined} values mark the best and second-best non-human results, respectively.
}
\begin{tabular*}{\textwidth}
{@{\extracolsep{\fill}}llcccccccc}
\toprule
\textbf{Method}
& \textbf{Setting/Backbone}
& \multicolumn{4}{c}{\textbf{MuMA-ToM}}
& \multicolumn{3}{c}{\textbf{MMToM-QA}}
& \multicolumn{1}{c}{\textbf{Hi-ToM}} \\
\cmidrule(lr){3-6}
\cmidrule(lr){7-9}
\cmidrule(lr){10-10}

&
& \textbf{Belief}
& \textbf{Social}
& \textbf{BoG}
& \textbf{Avg.}
& \textbf{Belief}
& \textbf{Goal}
& \textbf{Avg.}
& \textbf{Avg.} \\

\midrule
\multicolumn{10}{l}{\emph{Human Performance}} \\
\midrule

Human
& --
& 98.90 & 94.40 & 87.10 & 93.47
& 97.50 & 88.50 & 93.00
& -- \\

\midrule
\multicolumn{10}{l}{\emph{General-purpose MLLMs}} \\
\midrule

GPT-4o
& \multirow[c]{4}{*}{Baseline}
& 85.33 & 57.00 & 48.33 & 63.55
& 55.67 & 32.33 & 44.00
& 50.00 \\

GPT-5.4
&
& 89.67 & 84.67 & 82.00 & 85.45
& \underline{98.00} & 49.67 & 73.84
& 87.00 \\



Qwen3-VL-8B
&
& 65.33 & 53.33 & 26.33 & 48.33
& 71.67 & 44.67 & 58.17
& 53.00 \\

InternVL3.5-8B
&
& 51.00 & 46.33 & 46.00 & 47.78
& 42.67 & 27.33 & 35.00
& 44.00 \\

\midrule
\multicolumn{10}{l}
{\emph{ToM-Specific Reasoning Prompting Methods}} \\
\midrule

SimToM (ACL'24)
& GPT-4o
& 54.60 & 43.50 & 44.80 & 47.63
& 75.67 & 26.33 & 51.00
& 71.00 \\

ThoughtTracing (COLM'25)
& GPT-4o (Text-Only)
& -- & -- & -- & --
& 78.40 & 37.80 & 58.10
& -- \\


\midrule
\multicolumn{10}{l}{\emph{State-Tracking Methods}} \\
\midrule

DWM (EMNLP'24)
& GPT-4o (Text-Only)
& 76.70 & 53.30 & 60.00 & 63.33
& 46.00 & 28.50 & 37.25
& 58.00 \\

PDDL-Mind
& GPT-4o (Text-Only)
& 88.70 & \textbf{92.00} & \underline{83.30} & 88.00
& 91.00 & 85.70 & 88.35
& -- \\

\midrule
\multicolumn{10}{l}
{\emph{Model-Based Mental-State Inference}} \\
\midrule

BIP-ALM (ACL'24)
& GPT-4o
& 41.20 & 34.10 & 30.60 & 35.30
& 64.33 & 48.00 & 56.17
& 14.50 \\

LIMP (AAAI'25)
& GPT-4o
& 93.40 & 67.70 & 68.70 & 76.60
& 60.00 & 50.67 & 55.33
& 6.50 \\

\multirow[c]{3}{*}{AutoToM (NeurIPS'25)}
& GPT-4o
& 88.33 & 77.00 & 79.00 & 81.44
& 96.67 & 69.33 & 83.00
& 72.50 \\

& Qwen3-VL-8B
& 41.67 & 41.67 & 30.00 & 37.78
& 79.33 & 50.67 & 65.00
& 39.00 \\

& InternVL3.5-8B
& 25.67 & 34.00 & 30.67 & 30.11
& 76.33 & 49.00 & 62.67
& 34.50 \\

\midrule
\multicolumn{10}{l}{\emph{Our Method}} \\
\midrule


\multirow[c]{2}{*}{
    \shortstack[l]{
        \textbf{PIP-ToM}\\
        \textbf{(Open-source Models)}
    }
}
& Qwen3-VL-8B
& 91.33 & 82.00 & 82.00 & 85.11
& 90.33 & 78.00 & 84.17
& 87.00 \\

& InternVL3.5-8B
& 93.67 & 83.67 & 74.33 & 83.89
& 93.33 & 81.33 & 87.33
& 79.50 \\

\midrule

\multirow[c]{2}{*}{
    \shortstack[l]{
        \textbf{PIP-ToM}\\
        \textbf{(Closed-source Models)}
    }
}
& GPT-4o
& \underline{96.67} & 84.33
& \textbf{96.67} & \underline{92.56}
& \textbf{99.67} & \underline{90.33}
& \underline{95.00}
& \underline{95.00} \\

& GPT-5.4
& \textbf{98.00} & \underline{89.00}
& \textbf{96.67} & \textbf{94.56}
& \textbf{99.67} & \textbf{91.33}
& \textbf{95.50}
& \textbf{95.50} \\

\bottomrule
\end{tabular*}
\label{tab:main_results}
\end{table*}

\subsection{Baselines}

We compare PIP-ToM against state-of-the-art baselines:
\begin{itemize}
\setlength{\itemsep}{1pt}
\setlength{\parskip}{0pt}
\setlength{\parsep}{0pt}

\item \textbf{General-purpose MLLMs:}
GPT-4o~\citep{openai2024gpt4o},
GPT-5.4~\citep{openai2026gpt54},
Qwen3-VL-8B~\citep{bai2025qwen3vltechnicalreport}, and
InternVL3.5-8B~\citep{wang2025internvl35advancingopensourcemultimodal}.

\item \textbf{ToM-specific Reasoning Prompting Methods:}
SimToM~\citep{Wilf2023ThinkTP} and
ThoughtTracing~\citep{Kim2025HypothesisDrivenTR}.

\item \textbf{State Tracking Methods:}
DWM~\citep{Huang2024ANO} and
PDDL-Mind~\citep{Zhu2026PDDLMindLL}.

\item \textbf{Model-based Inference:}
BIP-ALM~\citep{jin2024mmtom},
LIMP~\citep{Shi2024MuMA_ToM}, and
AutoToM~\citep{zhang2025autotom}.
\end{itemize}

To isolate the contribution of the reasoning framework from differences in
upstream visual processing, PIP-ToM and AutoToM~\citep{zhang2025autotom} use identical fused video--text event traces. This ensures that both methods receive the same
reconstructed episode information, so their performance differences are not
attributable to different visual encoders or event-extraction pipelines.

\begin{figure}[t]
\centering
\includegraphics[width=\columnwidth]{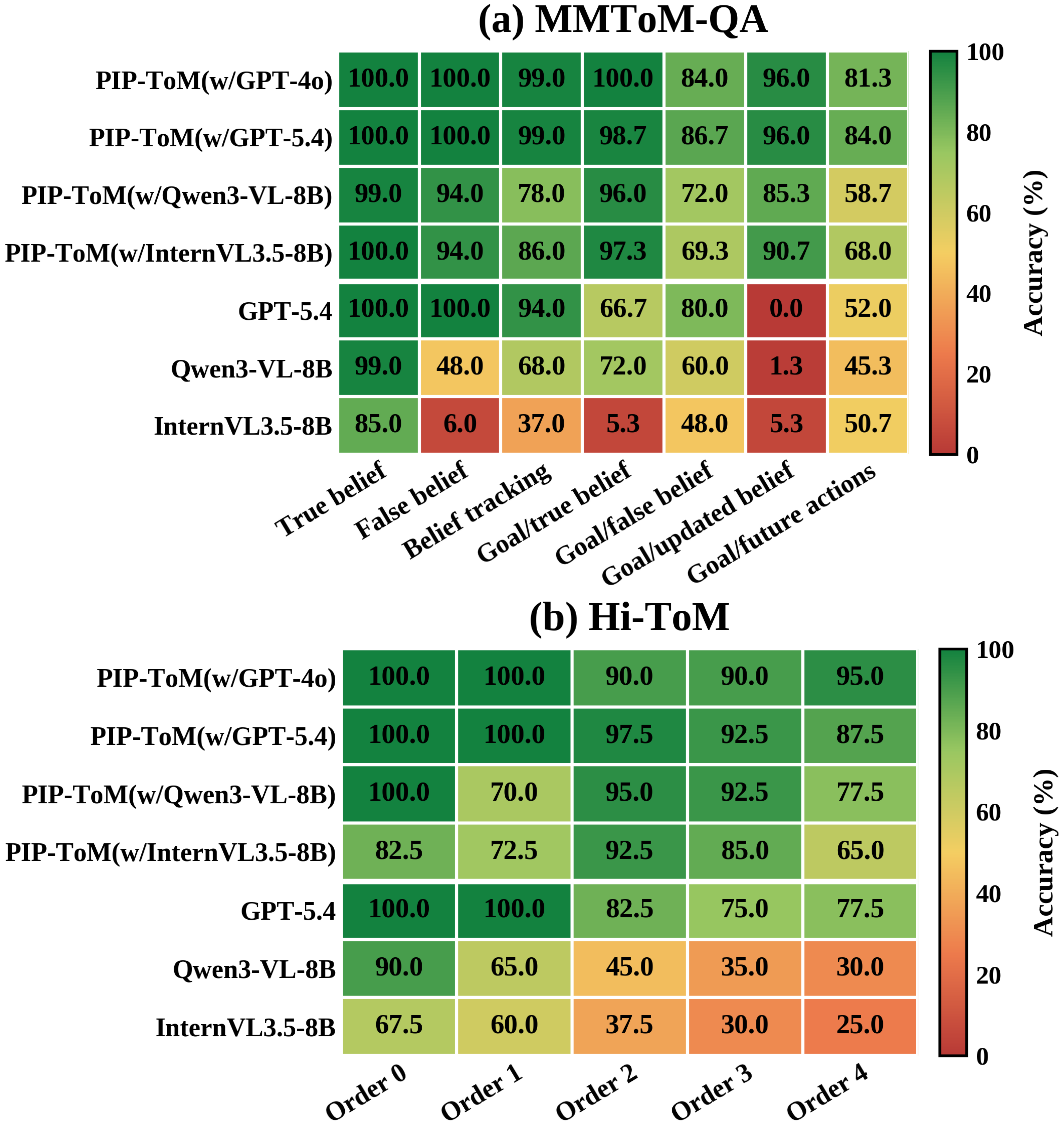}
\caption{Fine-grained accuracy across MMToM-QA belief and goal subtasks and
Hi-ToM reasoning orders 0--4.}
\label{fig:MMToM_HiToM_Accuracy}
\end{figure}

\subsection{Overall Performance}

\begin{figure*}[t]
\centering
\includegraphics[width=\textwidth]{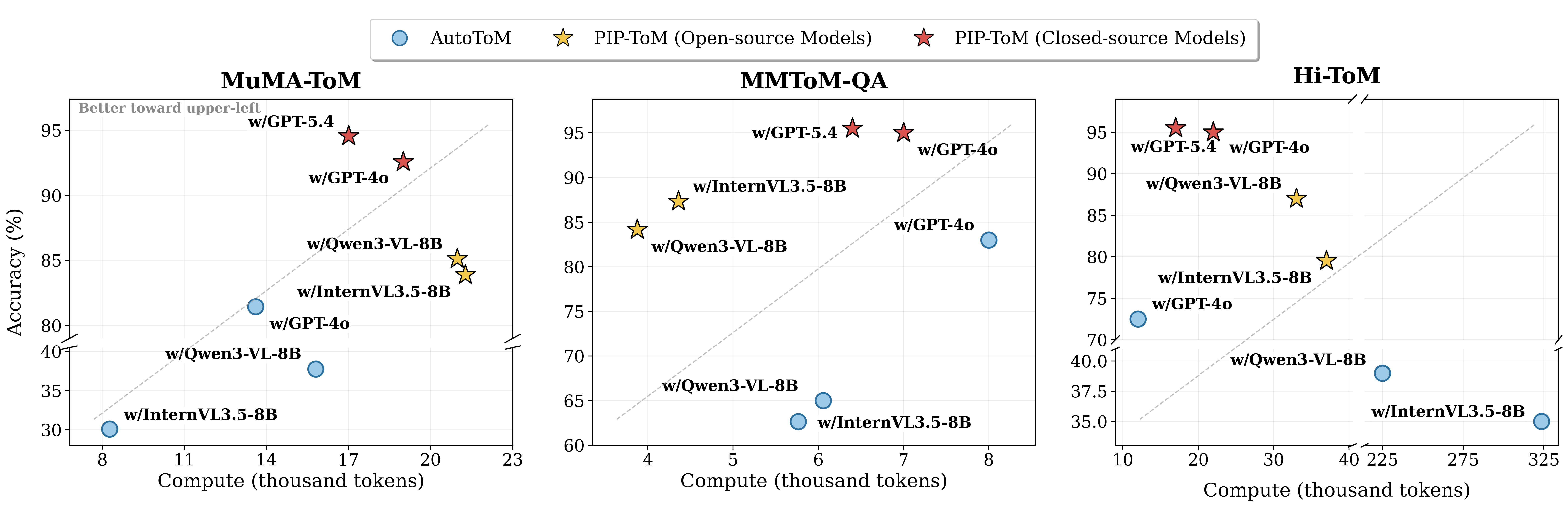}
\caption{
Averaged accuracy and inference-token consumption of PIP-ToM (stars) and AutoToM (circles) on three benchmarks.
}
\label{fig:acc-compute}
\end{figure*}

\subsubsection{Performance across Tasks and Reasoning Complexity}


Table~\ref{tab:main_results} summarizes performance across the three benchmarks. With GPT-4o, PIP-ToM achieves average accuracies of 92.56\%, 95.00\%, and 95.00\% on MuMA-ToM, MMToM-QA, and Hi-ToM, respectively, exceeding AutoToM by 11.12, 12.00, and 22.50 percentage points. On MuMA-ToM, the largest improvement occurs in Belief of Goal, where accuracy
increases from 79.00\% to 96.67\%. This gain extends beyond physical-state belief tracking to questions requiring one agent to attribute another agent's goal.

The improvements also extend to both open-source backbones. On MMToM-QA,
PIP-ToM improves over AutoToM by 19.17 percentage points with Qwen3-VL-8B
and 24.67 points with InternVL3.5-8B. Both backbones also improve on MuMA-ToM and Hi-ToM, suggesting that the benefits of PIP-ToM extend beyond a particular proprietary model. Nevertheless, task-level performance remains uneven: with GPT-4o, MuMA-ToM Social Goal accuracy reaches 84.33\%, below the 96.67\% achieved on both Belief and Belief of Goal, identifying social intention inference as a remaining challenge.

Figure~\ref{fig:MMToM_HiToM_Accuracy} provides a finer-grained view of task difficulty and reasoning depth. On MMToM-QA, substantial gains over direct prompting occur in belief-sensitive goal inference; for example,
Qwen3-VL-8B improves from 1.3\% to 85.3\% on Goal given updated belief.
Future-action-conditioned goal inference remains comparatively difficult across backbones. On Hi-ToM, PIP-ToM improves performance on nested belief questions, including order~4, where Qwen3-VL-8B and InternVL3.5-8B increase from 30.0\% to 77.5\% and from 25.0\% to 65.0\%, respectively, relative to direct prompting. These improvements show that the framework remains
useful for deeper attribution chains.

Furthermore, to assess sensitivity to random seeds, we evaluate PIP-ToM with InternVL3.5-8B using three different seeds (42, 1234, and 2024). The resulting standard errors are 0.13 percentage points on MuMA-ToM and 0.22 percentage points on MMToM-QA, indicating stable performance across the evaluated random seeds. Full results are provided in Supplementary Sec.~B.


\subsubsection{Accuracy-Compute Trade-off}


Figure~\ref{fig:acc-compute} compares accuracy and average inference-token consumption across the three benchmarks. Table~\ref{tab:token_dataset_summary} further reports the dataset-level costs of full methods and component-ablation variants.

The computational profile varies across backbones and task structures. MMToM-QA consistently incurs lower inference costs than MuMA-ToM and Hi-ToM, where social interactions or recursive belief attribution may require more extensive reasoning. With GPT-4o, PIP-ToM uses fewer tokens than AutoToM on MMToM-QA (7.00K vs. 8.00K), but more on MuMA-ToM (19.00K vs. 13.60K) and Hi-ToM (22.77K vs. 12.00K). For the open-source backbones, PIP-ToM substantially reduces token consumption on Hi-ToM, from 225.00K to 33.00K with Qwen3-VL-8B and from 323.42K to 37.00K with InternVL3.5-8B. On the multimodal benchmarks, PIP-ToM generally incurs higher costs on MuMA-ToM but lower costs on MMToM-QA.

The component ablations further reveal non-uniform computational effects. Removing either Perspective Storage or Claim-Aware Calibration reduces token consumption with GPT-4o on both multimodal benchmarks. In contrast, removing Perspective Storage with Qwen3-VL-8B increases the cost on MuMA-ToM from 20.97K to 35.20K, while producing little change on MMToM-QA. These results indicate that the computational impact of individual components depends on both the backbone and benchmark. Overall, the accuracy improvements of PIP-ToM are not driven by uniformly higher inference costs, although its computational efficiency remains task- and backbone-dependent.


\begin{table*}[t]
\centering
\small
\setlength{\tabcolsep}{6pt}
\renewcommand{\arraystretch}{1.15}
\caption{\textbf{Inference-token cost per question (K).}
(a) Comparison of AutoToM and PIP-ToM across different backbones and benchmarks.
(b) Inference-token costs of PIP-ToM component-ablation variants on MuMA-ToM and MMToM-QA.
Token consumption is averaged over completed inference instances, including both prompt and completion tokens across all model calls.}
\begin{tabular*}{\textwidth}{@{\extracolsep{\fill}}ccccccc@{}}
\toprule
\multicolumn{7}{c}{\textbf{(a) Full-method comparison}} \\
\addlinespace[3pt]

\multirow{2}{*}{\textbf{Backbone}}
& \multicolumn{2}{c}{\textbf{MuMA-ToM}}
& \multicolumn{2}{c}{\textbf{MMToM-QA}}
& \multicolumn{2}{c}{\textbf{Hi-ToM}} \\

\cmidrule(lr){2-3}
\cmidrule(lr){4-5}
\cmidrule(lr){6-7}

& \textbf{AutoToM}
& \textbf{PIP-ToM}
& \textbf{AutoToM}
& \textbf{PIP-ToM}
& \textbf{AutoToM}
& \textbf{PIP-ToM} \\

\midrule

GPT-4o
& 13.60 & 19.00
& 8.00 & 7.00
& 12.00 & 22.77 \\

GPT-5.4
& -- & 17.00
& -- & 6.40
& -- & 17.00 \\

Qwen3-VL-8B
& 15.80 & 20.97
& 6.06 & 3.88
& 225.00 & 33.00 \\

InternVL3.5-8B
& 8.27 & 21.27
& 5.76 & 4.36
& 323.42 & 37.00 \\

\bottomrule
\end{tabular*}

\medskip

\begin{tabular*}{\textwidth}{@{\extracolsep{\fill}}cccc@{}}
\toprule
\multicolumn{4}{c}{\textbf{(b) Component-ablation costs}} \\
\addlinespace[3pt]

\textbf{Setting}
& \textbf{Backbone}
& \textbf{MuMA-ToM}
& \textbf{MMToM-QA} \\

\midrule

\multirow{2}{*}{w/o Perspective Storage}
& GPT-4o
& 13.02 & 4.92 \\

& Qwen3-VL-8B
& 35.20 & 3.77 \\

\midrule

w/o Claim-Aware Calibration
& GPT-4o
& 11.96 & 5.94 \\

\bottomrule
\end{tabular*}

\label{tab:token_dataset_summary}

\end{table*}

\subsubsection{Statistical Significance Analysis}
We assess the reliability of PIP-ToM's gains over AutoToM using 3,400 question-level pairs across three benchmarks and two open-source backbones. Both methods use identical inputs, candidate answer ordering, backbone configurations, and random seeds. We estimate 95\% confidence intervals (CIs) for accuracy gains using cluster bootstrap~\cite{field2007bootstrapping} with 10,000 resamples, accounting for dependencies within episodes or stories. We resample 225 MuMA-ToM episodes, 134 MMToM-QA episodes, and 40 Hi-ToM stories. We also perform two-sided exact McNemar tests~\cite{mcnemar1947note} at the question level, with Holm--Bonferroni correction~\cite{holm1979simple} across the six comparisons.

\begin{table*}[t]
\centering
\caption{\textbf{Statistical significance of improvements over AutoToM.} Accuracy gains ($\Delta$) are in percentage points (pp). The 95\% CIs use cluster bootstrap with 10,000 resamples; $p$ and $p_{\mathrm{adj}}$ are the raw and Holm--Bonferroni-adjusted two-sided exact McNemar $p$-values.}
\label{tab:statistical_significance}
\small
\setlength{\tabcolsep}{5pt}
\renewcommand{\arraystretch}{1.15}
\resizebox{\textwidth}{!}{
\begin{tabular}{@{}llcccccc@{}}
\toprule
\multirow{2}{*}{\textbf{Backbone}} & \multirow{2}{*}{\textbf{Benchmark}} & \multicolumn{2}{c}{\textbf{Accuracy (\%)}} & \multirow{2}{*}{$\boldsymbol{\Delta}$ \textbf{(pp)}} & \multirow{2}{*}{\textbf{95\% CI}} & \multicolumn{2}{c}{\textbf{McNemar Test}} \\
\cmidrule(lr){3-4}\cmidrule(lr){7-8}
& & \textbf{AutoToM} & \textbf{PIP-ToM} & & & $\boldsymbol{p}$ & $\boldsymbol{p_{\mathrm{adj}}}$ \\
\midrule
\multirow{3}{*}{Qwen3-VL-8B}
& MuMA-ToM & 37.78 & \textbf{85.11} & +47.33 & [43.33, 51.44] & $1.73\times10^{-96}$ & $8.65\times10^{-96}$ \\
& MMToM-QA & 65.00 & \textbf{84.17} & +19.17 & [13.40, 24.88] & $5.38\times10^{-17}$ & $5.38\times10^{-17}$ \\
& Hi-ToM & 39.00 & \textbf{87.00} & +48.00 & [38.00, 58.00] & $1.25\times10^{-23}$ & $3.75\times10^{-23}$ \\
\midrule
\multirow{3}{*}{InternVL3.5-8B}
& MuMA-ToM & 30.11 & \textbf{83.89} & +53.78 & [49.78, 57.78] & $1.50\times10^{-115}$ & $9.00\times10^{-115}$ \\
& MMToM-QA & 62.67 & \textbf{87.33} & +24.67 & [20.20, 29.27] & $1.59\times10^{-29}$ & $6.36\times10^{-29}$ \\
& Hi-ToM & 34.50 & \textbf{79.50} & +45.00 & [35.00, 55.00] & $1.51\times10^{-18}$ & $3.02\times10^{-18}$ \\
\bottomrule
\end{tabular}
}
\end{table*}

As shown in Table~\ref{tab:statistical_significance}, PIP-ToM improves accuracy by 19.17--53.78 pp across all six comparisons. All cluster-bootstrap 95\% CIs exclude zero; even for the smallest gain (Qwen3-VL-8B on MMToM-QA), the lower bound is 13.40 pp. The gains on MuMA-ToM and Hi-ToM are substantial for both backbones.

All question-level McNemar tests remain significant after
Holm--Bonferroni correction ($p_{\mathrm{adj}}<0.001$).
These tests do not account for within-cluster dependence and are therefore complementary to the cluster-bootstrap intervals, all of which exclude zero.


\subsection{Counterfactual Evaluation of Evidence Admissibility}
\label{sec:counterfactual}

Existing ToM benchmarks primarily evaluate mental-state reasoning through question-answering accuracy, which does not explicitly assess whether predictions respond appropriately to changes in agent-accessible evidence. According to the Evidence Admissibility Principle, belief predictions should remain invariant to inaccessible changes while responding appropriately to changes in accessible, claim-relevant evidence. To examine these complementary properties, we construct a controlled counterfactual evaluation of perspective-sensitive ToM reasoning.

\begin{table*}[t]
\centering
\caption{\textbf{Counterfactual evaluation of evidence admissibility (\%).} All methods use GPT-4o. \textbf{World-State Accuracy} (World Acc.) and \textbf{Belief Accuracy} (Belief Acc.) measure the correctness of objective world-state and agent-specific belief predictions, respectively. \textbf{Correct Invariance} (CInv) requires correct and identical predictions across hidden conditions, while \textbf{Correct Switching} (CS) requires correct and distinct predictions across visible conditions. \textbf{Access Sensitivity} (AS) requires correct belief changes across hidden--visible pairs for the same relocation. \textbf{Quadruplet Accuracy} (Quad.) requires all four belief predictions within a group to be correct. CInv, CS, and Quad. are evaluated over 50 groups, and AS over 100 pairs.}
\label{tab:counterfactual}
\small
\setlength{\tabcolsep}{6pt}
\renewcommand{\arraystretch}{1.15}
\begin{tabular*}{\textwidth}{@{\extracolsep{\fill}}lcccccc@{}}
\toprule
\textbf{Method} & \textbf{World Acc.} $\uparrow$ & \textbf{Belief Acc.} $\uparrow$ & \textbf{CInv} $\uparrow$ & \textbf{CS} $\uparrow$ & \textbf{AS} $\uparrow$ & \textbf{Quad.} $\uparrow$ \\
\midrule
GPT-4o & \textbf{100.00} & 57.00 & 32.00 & 42.00 & 24.00 & 6.00 \\
AutoToM & 99.00 & 57.50 & 40.00 & 34.00 & 32.00 & 18.00 \\
PIP-ToM (w/o Perspective Storage) & \textbf{100.00} & 61.00 & 12.00 & \textbf{100.00} & 22.00 & 12.00 \\
\midrule
\textbf{PIP-ToM} & \textbf{100.00} & \textbf{98.50} & \textbf{98.00} & 98.00 & \textbf{97.00} & \textbf{96.00} \\
\bottomrule
\end{tabular*}
\end{table*}

\subsubsection{Experimental Setup.}
We adapt 50 distinct scenarios from Hi-ToM at the event-trace level. In each scenario, the target agent observes an object at location $A$ before it is relocated to either $B$ or $C$. By independently manipulating the relocation destination and the target agent's access to the relocation, we construct four matched conditions: $B_{\mathrm{hidden}}$, $C_{\mathrm{hidden}}$, $B_{\mathrm{visible}}$, and $C_{\mathrm{visible}}$. Within each group, the pre-intervention belief, background events, question wording, and candidate answer order are held fixed. To reduce template biases, we vary the surrounding event structures across groups and select alternative destinations with equivalent object-placement histories. Each condition includes a world-state question and a first-order belief question, yielding 400 questions in total. We compare Direct LLM, AutoToM, PIP-ToM without Perspective Storage, and Full PIP-ToM using GPT-4o as the common backbone. All methods receive identical complete event traces, questions, and ordered answer candidates.

\begin{figure}[t]
    \centering
    \includegraphics[width=\columnwidth]{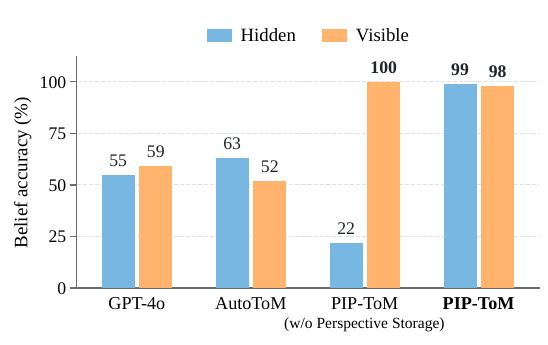}
    \caption{\textbf{Belief accuracy under hidden and visible conditions.} Removing Perspective Storage substantially degrades belief reasoning under hidden conditions, while preserving accuracy under visible conditions.}
    \label{fig:hidden_visible}
\end{figure}

\begin{figure}[t]
\centering
\includegraphics[width=\columnwidth]{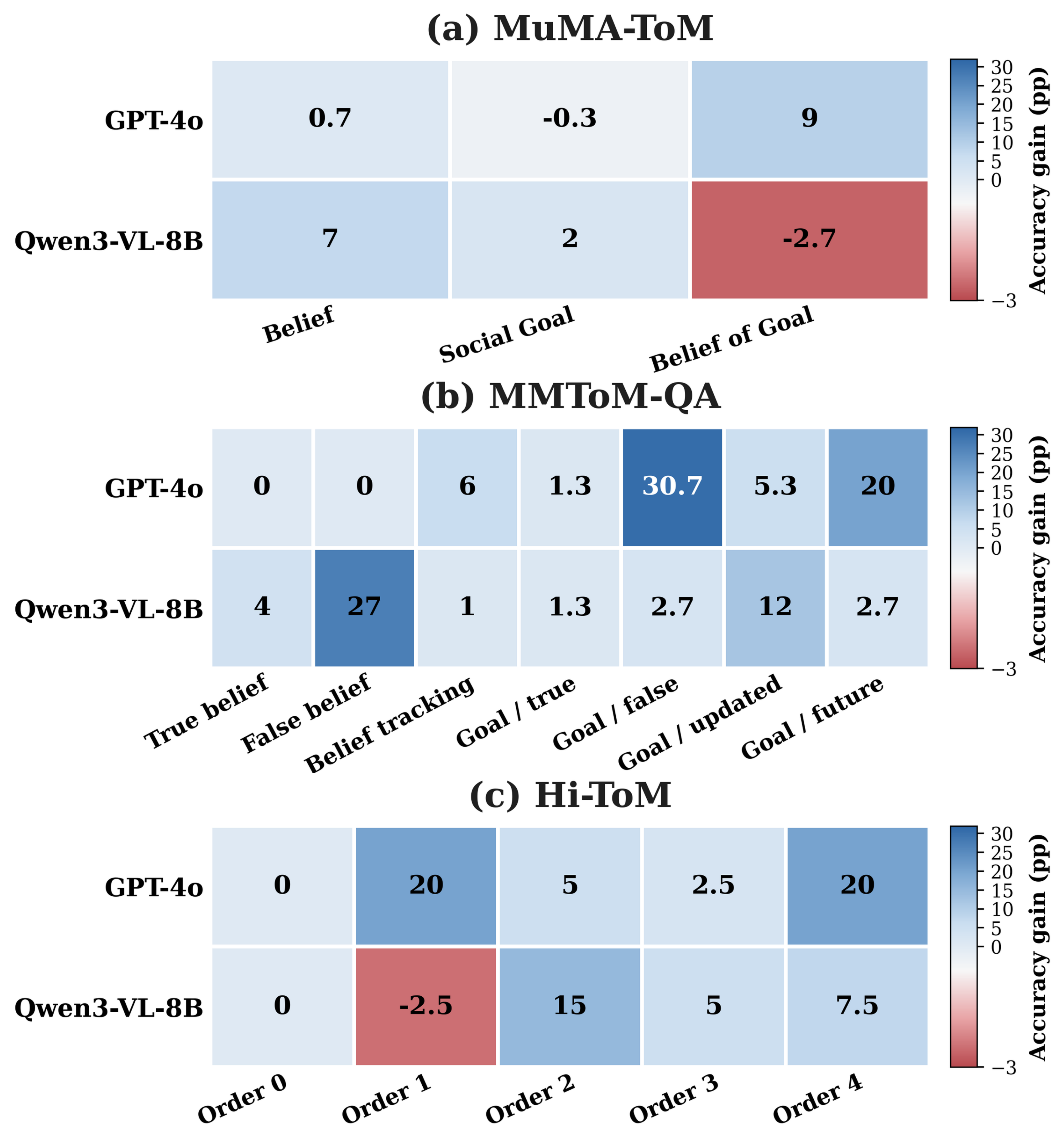}
\caption{Subtask-level accuracy differences (pp) between the Full model and w/o Perspective Storage.}
\label{fig:evidence_boundary}
\end{figure}

\subsubsection{Results and Analysis.}
As shown in Table~\ref{tab:counterfactual}, all methods achieve comparable world-state accuracy (99--100\%), yet exhibit substantially different belief reasoning behavior. PIP-ToM achieves 98\% CInv, 98\% CS, and 97\% AS, demonstrating its ability to preserve beliefs under inaccessible changes and appropriately update them in response to accessible evidence. Its quadruplet accuracy reaches 96\%, compared with 18\% for AutoToM, indicating more consistent reasoning across counterfactual conditions.

Removing Perspective Storage reduces CInv from 98\% to 12\% and AS from 97\% to 22\%, despite maintaining perfect world-state accuracy and CS. As further illustrated in Fig.~\ref{fig:hidden_visible}, the ablated model achieves 100\% belief accuracy under visible conditions but only 22\% under hidden conditions. Notably, all 78 incorrect belief predictions under hidden conditions match the objective world state, a pattern consistent with interference from inaccessible global information. In contrast, Full PIP-ToM achieves 99\% and 98\% accuracy under hidden and visible conditions, respectively.

These results provide controlled behavioral evidence that PIP-ToM promotes belief invariance under inaccessible changes and appropriate belief updates in response to accessible, claim-relevant evidence.

\subsection{Analysis by Research Questions}



We further analyze the effectiveness of PIP-ToM through three research questions: whether agent-specific evidence boundaries reduce interference from global information (\textbf{RQ1}), whether claim-aware evidence routing and calibration align admissible evidence with the mental-state variables it can support (\textbf{RQ2}), and whether perspective-conditioned evidence matching alone is sufficient across different mental-state reasoning tasks (\textbf{RQ3}).

\begin{table}[t]
\centering
\caption{\textbf{Component ablations with GPT-4o.}
The ablation results are averaged over two runs.
\textit{w/o Perspective Storage} uses the complete episode trace
during calibration, while \textit{w/o Claim-Aware Calibration}
removes the structured mental-state schema and route-specific calibration.}
\label{tab:ablation}

\scriptsize
\setlength{\tabcolsep}{0.5pt}
\renewcommand{\arraystretch}{1.10}

\begin{tabular*}{\columnwidth}{@{\extracolsep{\fill}}llccc@{}}
\toprule
\textbf{Benchmark}
& \textbf{Task}
& \textbf{Full}
& \textbf{\shortstack{w/o\\Perspective\\Storage}}
& \textbf{\shortstack{w/o\\Claim-Aware\\Calibration}} \\
\midrule
\multirow{4}{*}{MuMA-ToM}
& Belief & 96.67 & 96.00 & 86.83 \\
& Social & 84.33 & 84.67 & 67.67 \\
& BoG    & 96.67 & 87.67 & 86.33 \\
& Avg.   & 92.56 & 89.45 & 80.28 \\
\midrule
\multirow{3}{*}{MMToM-QA}
& Belief & 99.67 & 97.67 & 96.67 \\
& Goal   & 90.33 & 76.00 & 70.50 \\
& Avg.   & 95.00 & 86.84 & 83.59 \\
\bottomrule
\end{tabular*}
\end{table}

\subsubsection{RQ1: Perspective-Bounded Evidence}

\paragraph{Task-dependent effects of perspective storage.}

To test whether agent-specific boundaries limit interference from inaccessible global events, we compare the Full model with the \textit{w/o Perspective Storage} variant in Table~\ref{tab:ablation}. Removing the perspective boundary lowers average accuracy, particularly on MMToM-QA.

Figure~\ref{fig:evidence_boundary} further shows that this effect is selective rather than uniform. The largest gains from Perspective Storage occur on tasks in which the agent's information state may diverge from the objective episode state, including false-belief goal inference, belief updating, future-action prediction, cross-agent goal attribution, and higher-order belief reasoning. By contrast, performance changes little on true-belief and order-0 tasks, where agent-relative and global information largely align.


These differences can be understood by considering what the agent actually knows. For example, if an object is moved without the agent witnessing the relocation, its new location should not be used as direct evidence of the agent's belief. Perspective Storage preserves this distinction between what happens in the world and what the agent knows about it. Its benefits are therefore more pronounced in tasks involving belief updates, belief-dependent goals, and nested mental-state attribution.

\begin{table*}[t]
\centering
\caption{\textbf{Mechanism-level audit of perspective ablation errors.} The 121 cases correctly answered by the Full model but incorrectly answered by \textit{w/o Perspective Storage} are categorized into (A) error patterns consistent with inaccessible or global evidence interference and (B) other error patterns. Percentages are calculated over the 121 audited cases.}
\label{tab:perspective_audit}
\small
\setlength{\tabcolsep}{5pt}
\renewcommand{\arraystretch}{1.16}
\begin{tabular*}{\textwidth}{@{\extracolsep{\fill}}>{\raggedright\arraybackslash}p{0.29\textwidth}c>{\raggedright\arraybackslash}p{0.50\textwidth}@{}}
\toprule
\textbf{Error mechanism} & \textbf{Cases, n (\%)} & \textbf{Observed error pattern} \\
\midrule

\multicolumn{3}{@{}l}{\textbf{A. Inaccessible/global-evidence errors}\quad \textbf{(76 cases, 62.8\%)}} \\
\addlinespace[3pt]

Direct objective-state attraction & 22 (18.2\%) & The prediction matches an objective state unknown to the agent, after accounting for question polarity. \\
\addlinespace[2pt]

Relation-level interference & 37 (30.6\%) & Events unknown to the agent are used to explain actions or other observed events. \\
\addlinespace[2pt]

Candidate-grounding interference & 17 (14.0\%) & Global information leads to incorrect associations between candidate answers and objects, locations, or action targets. \\

\midrule

\multicolumn{3}{@{}l}{\textbf{B. Other error mechanisms}\quad \textbf{(45 cases, 37.2\%)}} \\
\addlinespace[3pt]

Residual non-matching errors & 16 (13.2\%) & No clear error pattern can be identified. \\
\addlinespace[2pt]

\textsc{Most}/\textsc{Least} polarity reversal & 6 (5.0\%) & The prediction matches the objective state even though the question asks for the least likely option. \\
\addlinespace[2pt]

Nondiagnostic objective state & 7 (5.8\%) & The objective state alone cannot distinguish the answer options. \\
\addlinespace[2pt]

Future-action or path reasoning & 15 (12.4\%) & Knowing the final state is insufficient to infer the queried goal or future action. \\
\addlinespace[2pt]

Invalid response & 1 (0.8\%) & The response does not identify a valid answer option. \\

\midrule
\textbf{Total} & \textbf{121 (100.0\%)} & \\
\bottomrule
\end{tabular*}
\end{table*}

\paragraph{Mechanism-level audit.}

To identify the mechanisms behind this improvement, we manually audit paired GPT-4o predictions from the Full and \textit{w/o Perspective Storage} configurations. Both configurations use identical inputs, candidate order, random seed, structured schema, calibration procedure, and Bayesian inference backbone. The comparison yields 151 Full-only cases, where the Full model is correct and the ablated model is incorrect.

We analyze 30 MuMA-ToM Social Goal cases separately because their candidates describe intentions or social effects, which cannot be directly aligned with objective physical states. The remaining 121 cases form the strictly comparable audit pool summarized in Table~\ref{tab:perspective_audit}. The audit distinguishes direct matches with inaccessible objective states from less direct interference in relational interpretation and candidate grounding, as well as other error patterns.

\paragraph{Patterns of global-information interference.}

Among the 121 audited cases, 76 (62.8\%) exhibit patterns consistent with interference from inaccessible or global evidence. Direct objective-state attraction accounts for 22 cases (18.2\%), where the ablated prediction matches an unobserved objective state rather than the queried agent's accessible belief.

The other 54 cases within this group (44.6\%) reveal two less direct forms of interference. In 37 relation-level cases, global information influences how observed actions are connected to state transitions or other agents' behavior. For example, a later outcome unavailable to the queried agent may be used to interpret that agent's earlier action. In 17 candidate-grounding cases, global information alters which entity, location, action, or behavioral target is associated with a candidate mental-state claim.


These three error patterns illustrate different ways in which global information can interfere with mental-state inference. A model may directly attribute an unobserved state to an agent, use events outside the agent's perspective to explain its behavior, or incorrectly associate a candidate answer with an object or action target. The latter two patterns are particularly important because global-information interference can occur even when the final prediction does not explicitly match the objective world state.

The remaining 45 cases (37.2\%) exhibit other error patterns, including question-polarity errors, nondiagnostic objective states, and future-action or path reasoning difficulties. Thus, direct objective-state attraction and other forms of global-information interference provide important but not exhaustive explanations for the errors introduced by removing Perspective Storage.

\paragraph{Task-selective objective-state attraction.}

Although direct objective-state attraction accounts for only 18.2\% of the strictly comparable audit pool, its occurrence is highly task-dependent. All 22 direction-consistent cases are concentrated in MMToM-QA belief tracking (6/6), updated-belief goal inference (4/6), and Hi-ToM belief reasoning (12/26). Together, these three settings account for 22 of 38 audited Full-only cases (57.9\%), whereas no direction-consistent objective-state matches are identified among the remaining 83 comparable cases. This retrospective distribution is consistent with the importance of evidence boundaries in tasks involving temporal belief updates and nested mental-state attribution.

The three settings impose different but related information-access requirements. Belief tracking requires preserving previously acquired information when subsequent world-state changes are inaccessible to the agent. Updated-belief goal inference additionally requires determining whether the agent has received evidence that warrants a belief update before evaluating its goal. Recursive belief reasoning requires maintaining information ownership across multiple agents and attribution levels.

The separately examined Social Goal cases reveal a different pattern. Among the 30 cases, 12 ablated predictions match the observed social outcome, but all occur under \textsc{Least} questions. Since an observed social outcome does not necessarily establish the speaker's underlying intention, these matches cannot be interpreted as direct objective-state attraction. Instead, they highlight the need to distinguish outcome salience, mental-state attribution, and question polarity when analyzing social-goal errors.

These findings clarify the role of Agent-Specific Perspective Storage beyond preventing direct objective-state attraction. By restricting the agent-scoped evidence view to observations, actions, utterances, and memories that satisfy the agent's access constraints, the module also limits the influence of inaccessible global information on relational interpretation and candidate grounding. Its contribution therefore extends from filtering inaccessible states to constraining the evidential basis on which candidate mental-state claims are evaluated.

\rqconclusion{Takeaway 1: Perspective-bounded evidence reduces global-information interference in perspective-sensitive ToM reasoning, extending beyond direct objective-state attraction to relational interpretation and candidate grounding.}

\subsubsection{RQ2: Claim-Aware Evidence Calibration}
\label{sec:rq2}
\paragraph{Task-dependent effects of claim-aware calibration.}

To examine the role of claim-aware evidence calibration, we compare the Full model with the \textit{w/o Claim-Aware Calibration} variant. As shown in Table~\ref{tab:ablation}, removing this component causes the largest overall performance decline among the evaluated ablations. Accuracy drops from 84.33\% to 67.67\% on MuMA-ToM Social Goal and from 90.33\% to 70.50\% on MMToM-QA Goal, with smaller reductions on explicit belief questions.

The relative effects of the two components further reveal different evidential requirements across mental-state types. On MuMA-ToM Social Goal, removing Perspective Storage produces little change in accuracy, whereas removing Claim-Aware Calibration causes a decline of 16.67 percentage points. This contrast is consistent with the nature of social-goal reasoning: access to an utterance alone does not determine the speaker's intention. Its evidential role depends on the relationship between the communicated content, the speaker's belief, and the anticipated effect on another agent.

In contrast, MMToM-QA Goal and MuMA-ToM Belief of Goal exhibit substantial performance declines under both ablations. On MMToM-QA Goal, removing Perspective Storage and Claim-Aware Calibration reduces accuracy by 14.33 and 19.83 percentage points, respectively. On MuMA-ToM Belief of Goal, the corresponding reductions are 9.00 and 10.33 percentage points. These tasks require not only identifying which information was available to the relevant agent, but also determining how that information supports a behavioral or recursively attributed hypothesis. The results highlight the complementary roles of evidence accessibility and claim-specific relevance in mental-state inference.

\begin{table}[t]
\centering
\footnotesize
\setlength{\tabcolsep}{3pt}
\renewcommand{\arraystretch}{1.05}
\caption{\textbf{Factor-scale robustness with GPT-5.4.}
Accuracy (\%) under different calibration strengths.
Parentheses indicate changes (percentage points) relative to
$\alpha=1.0$, computed before rounding.}
\label{tab:factor_scale}
\begin{tabular*}{\columnwidth}{@{\extracolsep{\fill}}lccc@{}}
\toprule
\textbf{Task}
& $\boldsymbol{\alpha=1.0}$
& \shortstack{\textbf{Weak}\\$(\alpha=0.75)$}
& \shortstack{\textbf{Strong}\\$(\alpha=1.25)$} \\
\midrule
\multicolumn{4}{@{}l}{\textit{MuMA-ToM}} \\
\cmidrule{1-4}
Belief         & 98.00 & 95.00 ($-3.00$) & 98.00 ($0.00$) \\
Social Goal    & 89.00 & 88.00 ($-1.00$) & 89.67 ($+0.67$) \\
Belief of Goal & 96.67 & 96.33 ($-0.33$) & 98.00 ($+1.33$) \\
Avg.           & 94.56 & 93.11 ($-1.44$) & 95.22 ($+0.67$) \\
\midrule
\multicolumn{4}{@{}l}{\textit{MMToM-QA}} \\
\cmidrule{1-4}
Belief & 99.67 & 99.67 ($0.00$)  & 99.67 ($0.00$) \\
Goal   & 91.33 & 87.33 ($-4.00$) & 91.67 ($+0.33$) \\
Avg.   & 95.50 & 93.50 ($-2.00$) & 95.67 ($+0.17$) \\
\bottomrule
\end{tabular*}
\end{table}

\paragraph{Calibration-strength sensitivity.}

Beyond determining which evidence relations are relevant to a mental-state claim, claim-aware calibration controls how strongly these relations influence the underlying model's hypothesis evaluation. To examine sensitivity to calibration strength, we globally rescale all active factors with $\alpha\in\{0.75,1.0,1.25\}$ using GPT-5.4:
\[
\Phi_{\alpha}=\Phi^{\alpha}.
\]

This transformation preserves the direction of each evidence adjustment while modifying its magnitude. Specifically, $\alpha<1$ moves non-neutral factors toward $1$, whereas $\alpha>1$ strengthens both supporting and suppressing adjustments. Neutral factors remain unchanged. The parser, route activations, base likelihoods, candidate order, and random seed are fixed across all settings. The $\alpha=1.0$ reference corresponds to the GPT-5.4 configuration reported in Table~\ref{tab:main_results}.

As shown in Table~\ref{tab:factor_scale}, benchmark-level performance remains relatively stable across the tested scales. We measure sensitivity as the mean absolute change in benchmark accuracy across the two perturbed settings relative to $\alpha=1.0$. The resulting deviations are 1.06 percentage points on MuMA-ToM and 1.08 percentage points on MMToM-QA, indicating limited aggregate sensitivity to moderate global rescaling.

\paragraph{Asymmetric effects of calibration strength.}

Despite the relatively small aggregate deviations, weakening and strengthening calibration produce different task-level effects. Weakening calibration reduces accuracy on four of the five evaluated subtasks, with the largest decline observed on MMToM-QA Goal ($-4.00$ percentage points). MuMA-ToM Belief also exhibits a noticeable reduction, whereas MMToM-QA Belief remains unchanged across all three settings.

This pattern suggests that the influence of admissible evidence during model-based inference matters in addition to identifying the appropriate evidence relations. When competing hypotheses receive similar or conflicting support from the underlying model, weaker calibration may limit the ability of structured evidence to influence their relative scores. This is particularly relevant to goal inference, where candidate hypotheses must be evaluated jointly with the agent's information state and observed behavior.

Increasing calibration strength, however, produces comparatively small and task-dependent changes. The largest improvement is observed on MuMA-ToM Belief of Goal ($+1.33$ percentage points), while both explicit-belief tasks remain nearly unchanged. Within the evaluated range, these results indicate that effective claim-aware inference benefits from sufficiently influential evidence adjustments, while further strengthening those adjustments does not necessarily yield comparable improvements.

These component ablations and factor-scale analysis highlight two complementary aspects of claim-aware calibration: the semantic alignment of evidence with the queried mental state and the influence of that evidence during subsequent model-based hypothesis evaluation.

\rqconclusion{Takeaway 2: Evidence accessibility alone is insufficient; claim-aware calibration aligns evidence with the queried mental state and enables it to effectively influence goal and social reasoning.}

\begin{table*}[t]
\centering
\small
\setlength{\tabcolsep}{3pt}
\renewcommand{\arraystretch}{1.15}
\caption{\textbf{Direct-posterior diagnostic with GPT-4o.} Coverage (\%) measures the proportion of questions admitting direct posterior construction. Variant and Full denote the overall accuracy of the hybrid variant and full PIP-ToM, respectively, with $\Delta$ reporting their difference in percentage points. V$\to$F reports paired subset accuracy (\%), while W2C/C2W counts wrong-to-correct and correct-to-wrong transitions.}
\resizebox{\textwidth}{!}{%
\begin{tabular}{@{}llcccccccc@{}}
\toprule
& & & \multicolumn{3}{c}{\textbf{Overall (\%)}} & \multicolumn{2}{c}{\textbf{Direct-covered}} & \multicolumn{2}{c}{\textbf{Fallback}} \\
\cmidrule(lr){4-6}
\cmidrule(lr){7-8}
\cmidrule(lr){9-10}
\textbf{Benchmark} & \textbf{Task} & \textbf{Coverage} & \textbf{Variant} & \textbf{Full} & $\boldsymbol{\Delta}$ & \textbf{V$\to$F} & \textbf{W2C/C2W} & \textbf{V$\to$F} & \textbf{W2C/C2W} \\
\midrule
\multirow{3}{*}{MuMA-ToM}
& Belief & 66.33 {\scriptsize (199/300)} & 92.67 & 96.67 & $+4.00$ & $100.00\!\to\!96.98$ & 0 / 6 & $78.22\!\to\!96.04$ & 21 / 3 \\
& Social Goal & 11.33 {\scriptsize (34/300)} & 78.00 & 84.33 & $+6.33$ & $85.29\!\to\!82.35$ & 2 / 3 & $77.07\!\to\!84.59$ & 48 / 28 \\
& Belief of Goal & 100.00 {\scriptsize (300/300)} & 76.00 & 96.67 & $\mathbf{+20.67}$ & $76.00\!\to\!96.67$ & 67 / 5 & -- & -- \\
\midrule
\multirow{2}{*}{MMToM-QA}
& Belief & 96.67 {\scriptsize (290/300)} & 99.67 & 99.67 & $0.00$ & $100.00\!\to\!100.00$ & 0 / 0 & $90.00\!\to\!90.00$ & 0 / 0 \\
& Goal & 61.67 {\scriptsize (185/300)} & 73.00 & 90.33 & $\mathbf{+17.33}$ & $65.95\!\to\!88.11$ & 50 / 9 & $84.35\!\to\!93.91$ & 16 / 5 \\
\bottomrule
\end{tabular}%
}

\label{tab:direct_posterior_diagnostic}
\end{table*}

\subsubsection{RQ3: Matching-Inference Complementarity}

\paragraph{Direct-posterior diagnostic.}

To investigate how perspective-conditioned evidence matching and model-based inference complement each other in mental-state reasoning, we evaluate a direct-posterior variant of PIP-ToM. When the activated evidence routes provide candidate-specific structured matching scores, the variant normalizes these scores directly to construct the answer posterior. Otherwise, it falls back to the uncalibrated Bayesian inference backbone.

This diagnostic examines matching--inference complementarity in terms of both the applicability and inferential capability of structured matching. Coverage measures the proportion of questions admitting direct posterior construction. Through paired comparisons on identical questions within the direct-covered and fallback subsets, we further examine how full PIP-ToM exploits available matching information and supports mental-state inference when direct matching is unavailable.

\paragraph{Direct matching effectiveness and inference complementarity.}

As shown in Table~\ref{tab:direct_posterior_diagnostic}, structured evidence matching is particularly effective for explicit belief inference. Direct posterior construction covers 66.33\% and 96.67\% of the Belief questions on MuMA-ToM and MMToM-QA, respectively, achieving 100.00\% accuracy on both covered subsets. These results demonstrate the effectiveness of perspective-conditioned matching in establishing relations between accessible evidence and candidate beliefs. When an agent's observations or source-preserving memories provide discriminative relations among candidate states, structured matching can effectively resolve the queried belief.

The paired comparison further reveals the complementary role of full inference. On MuMA-ToM Belief, full PIP-ToM improves fallback accuracy from 78.22\% to 96.04\%, correcting 21 predictions and yielding a net gain of 18 correct predictions. Meanwhile, it maintains high accuracy of 96.98\% and 100.00\% on the direct-covered belief subsets of MuMA-ToM and MMToM-QA, respectively. These results demonstrate that full PIP-ToM retains the advantages of structured matching while effectively addressing belief questions for which direct posterior construction is unavailable.

This complementarity is also evident on MuMA-ToM Social Goal, where the indirect relationship between observable behavior and latent social intentions limits direct coverage to 11.33\%. Full PIP-ToM corrects 48 predictions within the fallback subset, yielding a net gain of 20 correct predictions. This further demonstrates its ability to support latent mental-state inference when direct matching relations are limited.

\paragraph{From evidence matching to complex mental-state inference.}

Beyond compensating for limited direct matching coverage, full inference further exploits available structured matching information to support more complex mental-state reasoning.

This role is particularly evident on MuMA-ToM Belief of Goal. Direct posterior construction covers all 300 questions, demonstrating the broad availability of candidate-specific matching information. Building upon this information, full PIP-ToM improves accuracy from 76.00\% to 96.67\%, correcting 67 predictions and yielding a net gain of 62 correct predictions, corresponding to an improvement of 20.67 percentage points.

A similar pattern emerges on MMToM-QA Goal. Within the same 185 direct-covered questions, full PIP-ToM improves accuracy from 65.95\% to 88.11\%, correcting 50 predictions. It also improves fallback accuracy from 84.35\% to 93.91\%, resulting in an overall accuracy of 90.33\%.

These results reveal the complementary roles of structured matching and model-based inference in complex mental-state attribution. Goal and belief-of-goal reasoning require interpreting observed behavior in relation to competing latent goals and, in recursive settings, what the attributing agent could observe and infer about another agent's goal. Structured matching establishes candidate-specific compatibility with accessible evidence, while model-based inference integrates this compatibility into the dependencies among observations, behavior, and latent mental states.

These findings highlight the value of integrating perspective-conditioned evidence matching into model-based mental-state inference in PIP-ToM. The structured matching relations provide discriminative evidence for explicit belief claims, while the complete framework extends inference beyond directly matched cases and further exploits available matching information to reason about latent and recursively attributed mental states.

\rqconclusion{Takeaway 3: Evidence availability does not guarantee inference sufficiency; evidence-calibrated Bayesian reasoning integrates evidence into latent mental-state dependencies, enabling more effective goal and recursive mental-state inference.}

\subsection{Qualitative Case Studies}

\begin{figure*}[t]
\centering
\includegraphics[width=0.94\textwidth]{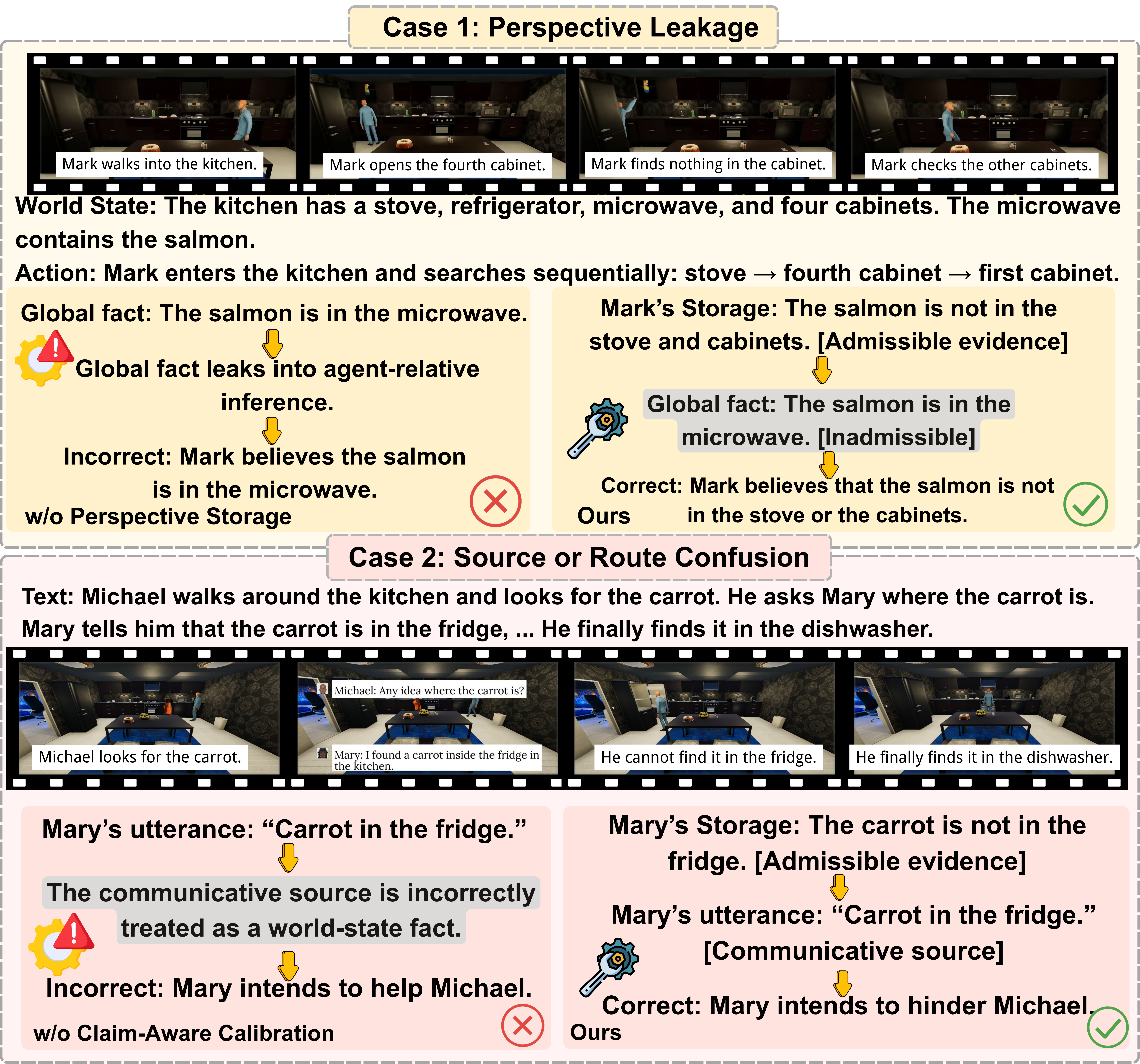}
\caption{
Qualitative effects of the two components.
Case 1 shows that perspective storage prevents inaccessible global facts from influencing inference.
Case 2 shows that claim-aware routing prevents accessible evidence from being misused.
}
\label{fig:qualitative_cases}
\end{figure*}

Figure~\ref{fig:qualitative_cases} distinguishes two questions that must be
resolved before evidence can contribute to mental-state inference: whether it
is accessible to the relevant agent, and whether its source and semantic role
allow it to support the queried mental-state claim. Agent-Specific Perspective
Storage addresses the first question, whereas Claim-Aware Calibration
addresses the second.

\subsubsection{Case 1: Perspective Leakage.}
The complete episode reveals that the salmon is inside the microwave. Mark,
however, searches only the stove and several cabinets and does not observe the
microwave's contents. Without Agent-Specific Perspective Storage, the globally
available location may be treated as part of Mark's information state, leading
to the incorrect conclusion that he believes the salmon is in the microwave.
PIP-ToM instead constructs an agent-specific perspective store from Mark's
observations and actions. His failed searches provide admissible evidence that
the salmon is not in the inspected locations, while its actual location
remains inaccessible and is excluded from belief inference. This case
demonstrates that inaccessible information should be treated as missing
evidence rather than as evidence available to the agent.

\subsubsection{Case 2: Source or Route Confusion.}
Mary tells Michael that the carrot is in the fridge, although her own
information indicates that it is not there. Without Claim-Aware Calibration,
the utterance may be flattened into the event trace and treated as an
objective statement about the carrot's location, making the instruction
appear helpful. PIP-ToM instead preserves the utterance as a source-tagged
communicative act and evaluates it relative to Mary's information state and
its expected effect on Michael's search. Because the communicated location
conflicts with what Mary knows and directs Michael toward an incorrect
location, the evidence supports a hindering rather than a helping intention.
The error therefore arises not from insufficient context but from assigning
accessible evidence to the wrong semantic role.

Together, the two cases show that evidence accessibility and evidence
diagnosticity impose complementary constraints. Agent-Specific Perspective
Storage prevents objective but agent-inaccessible events from entering the
inference process, whereas Claim-Aware Calibration prevents accessible
observations, actions, and utterances from supporting claims for which they
are not semantically appropriate. PIP-ToM therefore improves inference by
controlling how existing evidence is admitted and interpreted rather than by
introducing additional episode context.

\section{Conclusion}
\label{sec:conclusion}

In this work, we formulate evidence admissibility as a central
challenge in multimodal Theory of Mind and introduce PIP-ToM
to address this challenge through source-preserving perspective
storage, structured mental-state claims, and claim-aware evidence
calibration. By integrating these components into model-based
inference, PIP-ToM explicitly connects character-relative evidence
with the evaluation of competing mental-state hypotheses.

Extensive experiments and controlled event-trace counterfactual
evaluations demonstrate the effectiveness of PIP-ToM and highlight
the complementary roles of evidence accessibility, claim-dependent
relevance, and model-based inference. These findings underscore
the importance of understanding who had access to evidence,
how it was acquired, and what mental-state claims it can support.
We hope this evidence-centered perspective will facilitate
the development of more reliable and interpretable multimodal
social intelligence.

\section{Limitations}

\subsection{Dependence on reconstructed evidence.}
PIP-ToM relies on temporally ordered fused event traces. Errors in multimodal perception and ambiguous information-access cues may affect downstream mental-state reasoning, and robustness to such errors remains unverified.

\subsection{Task-aware schema and calibration.}
The structured schema and calibration rely on task-aware rules. Extending PIP-ToM to new mental-state variables may require additional schema fields, evidence routes, and calibration factors.

\subsection{Token consumption and scalability.}
Longer interaction histories and additional agents may increase token consumption due to larger perspective stores and more complex recursive attribution. Scalability to such interactions remains unverified.

\backmatter

\section*{Statements and Declarations}

\subsection*{Funding}
This work was supported in part by the National Natural Science Foundation of China under Grant~62406126.

\subsection*{Competing Interests}
A patent application related to the method described in this manuscript is pending. The authors declare no other competing interests.

\subsection*{Data Availability}
The benchmarks used in this study, including MuMA-ToM, MMToM-QA, and Hi-ToM, are publicly available from their respective official repositories. The counterfactual evaluation scenarios constructed in this study will be made publicly available upon publication.

\subsection*{Code Availability}
The source code and implementation details will be made publicly available upon publication.

\bibliography{main}

\end{document}


\maketitle

\begin{center}
\textit{Supplementary Information for the International Journal of Computer Vision}
\end{center}

\noindent
The supplementary material provides the implementation details omitted from
the main paper because of space limitations. It first specifies the
information-extraction interface, prompts, Agent-Specific Perspective Storage, route
activation, and decision procedure used by \pip{}. It then reports the benchmark
and baseline settings together with additional ablation and robustness
experiments. Finally, it describes the procedure used for the manual audit of
perspective-boundary errors reported in the main paper.

\medskip
\begingroup
\small
\DirectoryLine{0em}{A.}{jump-method}{\textbf{\pip{} Implementation Details}}
\DirectoryLine{1.4em}{A.1}{jump-info}{Information Extraction and Perspective Construction}
\DirectoryLine{2.8em}{A.1.1}{jump-prompts}{Prompts Used in \pip{}}
\DirectoryLine{2.8em}{A.1.2}{jump-schema}{Shared Schema and Task-Aware Instantiation}
\DirectoryLine{2.8em}{A.1.3}{jump-store}{Agent-Specific Perspective Storage}
\DirectoryLine{2.8em}{A.1.4}{jump-input}{Input Construction}
\DirectoryLine{1.4em}{A.2}{jump-routes}{Route Activation, Aggregation, and Factor Ranges}
\DirectoryLine{1.4em}{A.3}{jump-decision}{Task Instantiations and Evidence Handling}
\DirectoryLine{0em}{B.}{jump-experiments}{\textbf{Additional Results and Experimental Details}}
\DirectoryLine{1.4em}{B.1}{jump-baselines}{Baseline Implementation Details}
\DirectoryLine{1.4em}{B.2}{jump-benchmarks}{Benchmark Details}
\DirectoryLine{1.4em}{B.3}{jump-additional-experiments}{Additional Experiments}
\DirectoryLine{2.8em}{B.3.1}{jump-detailed-experimental-results}{Detailed Experimental Results}
\DirectoryLine{2.8em}{B.3.2}{jump-token-consumption}{Token Consumption}
\DirectoryLine{2.8em}{B.3.3}{jump-random-seed-robustness}{Robustness to Random Seeds}
\DirectoryLine{0em}{C.}{jump-audit}
{\textbf{Manual Audit of Perspective-Boundary Errors}}
\DirectoryLine{1.4em}{C.1}{jump-audit-scope}{Audit Scope and Label Hierarchy}
\DirectoryLine{1.4em}{C.2}{jump-audit-procedure}{Audit Procedure}
\endgroup
\vspace{0.5em}

\LinkTarget{jump-method}
\section{\pip{} Implementation Details}
\label{sec:method_details}

\LinkTarget{jump-info}
\subsection{Information Extraction and Perspective Construction}
\label{sec:parsing_prompts}

\paragraph{Hybrid parsing interface.}
The parser is executed once per question before probabilistic inference. A
fixed-prompt backbone identifies the shared query fields and normalizes
candidate wording. Task adapters retain the benchmark-native question,
candidate, and event semantics while constructing the latent variables needed
for the corresponding question family. They may select the inferred variable
and expose the structured relations used by the applicable evidence routes,
but they do not receive the ground-truth answer or assign the final candidate scores.
The parser returns
\[
R_q=\langle \tau,i,j,x,\rho\rangle,\qquad
Z_k=\langle R_q,\chi_k,\epsilon_k\rangle,
\]
where $\tau$ is the mental-state type, $i$ is the perspective holder, $j$ is
the target agent when one is present, $x$ is the queried entity, $\rho$ is the
question operator or relation, $\chi_k$ is the candidate value, and
$\epsilon_k$ is an optional behavioral or social effect. Prompts request
short structured outputs rather than free-form rationales.

\LinkTarget{jump-prompts}
\subsubsection{Prompts Used in \pip{}}
\label{sec:fixed_prompts}

The following prompts are shared across backbones. Bracketed fields are
replaced by the current example. Capitalization and output constraints are
kept fixed across all evaluated datasets.

\begin{PromptBox}{Identifying the perspective holder}
Find the name of the character that we need to infer about in the question
and choices. Only output the name. Do not answer the question.\\
{[Question]}\\
{[Choices]}\\
Character name:
\end{PromptBox}

\begin{PromptBox}{Identifying all agents}
Extract the names of all the characters from the story and question.
Provide only the names or roles, without any additional information.
Do not answer the question.\\
Your response should be a list containing the names, like
['name1', 'name2'].\\
{[Story]}\\
Response:
\end{PromptBox}

\begin{PromptBox}{Extracting an explicit question assumption}
If there is any assumed information in the question given (a conditional
clause starting with specific words like 'if' is contained), rewrite it as
a declarative sentence. Do not include any questions in the extra
information. Do not make up details for the information. Use the original
wording.\\
Otherwise, output 'NONE'.\\
Question: [Question]\\
Extra Information:
\end{PromptBox}

\begin{PromptBox}{Normalizing candidate answers}
Please write all choices as complete sentences based on the question,
including all the subjects.\\
If they are already complete sentences, output the choices as it was
following the format.\\
The output format should be ['aaa', 'bbb', ...]. Make sure to contain all
the choices.\\
{[Question]}\\
{[Choices]}\\
Make sure to contain all the choices.\\
Output:
\end{PromptBox}

\begin{PromptBox}{Identifying the queried mental-state variable}
Choose the variable that best summarizes the information of differences
that the choices contain.\\
Only output the variable.\\
Variables include: [Variables]\\
{[Choices]}\\
Variable:
\end{PromptBox}

\begin{PromptBox}{Extracting candidate entities}
Extract all the entities in the sentence: [Choice]\\
Entities include names of the agents, objects, locations, ...\\
The output format should be: ["aaa.", "bbb.", ...]. Output only this list.
\end{PromptBox}

\begin{PromptBox}{Extracting the inferred agent's action sequence}
Extract the actions of [Inferred\_agent] in the story verbatim without
changing any of the original words, pluralizing the words, adding in
[Inferred\_agent] or any other name, replacing any of the words, replacing
pronouns with names or replacing any names with pronouns. Actions of
[Inferred\_agent] are defined as events that will change the world state,
e.g., [Inferred\_agent] moving to a new location is an action but
[Inferred\_agent] being at a location is not an action. If
[Inferred\_agent] say something, the whole sentence (with 'replied', 'said')
is seen as an action.\\

Do not change the names of any of the agents, if there is not a name and only
a pronoun then just leave the pronoun. There can be more than one agent or more
than just the inferred agent. If there are multiple actions in a sentence
then they should be extracted as one single action, without changing any of
the original words, such as pluralizing the words, replacing any of the
words, replacing pronouns with names or replacing any names with pronouns,
and do not add any words.\\

Do not insert actions or pronouns or other words that are not explicitly
stated in the text. Do not separate the objects in the same action. Do not
add any pronouns. Keep the commas, if any. Only actions that have already
occurred at the time can be considered clearly stated. Again, only extract
actions performed by [Inferred\_agent].\\

The output format should be: ["aaa.", "bbb.", ...]. Output only this list.\\
{[Story]}\\
Extraction:
\end{PromptBox}

\begin{PromptBox}{Normalizing the extracted action list}
{[Story]}\\
Given the list of the actions of [Inferred\_agent] that are clearly stated
in the story without changing the original wording: [bare\_actions].\\
If it is already in full sentences and grammatically correct do not change
anything and output the same thing as the input. Do not change the names of
any agent.\\
If it is not in full sentences or not grammatically correct, then follow the
following steps.\\
1. If [Inferred\_agent] is not mentioned in the action then rephrase it so
that it has [Inferred\_agent].\\
2. Make sure that it is grammatically correct but do not lose any
information.\\
3. If the object involved in [Inferred\_agent]'s action is not clearly
described and for example just described as "it" then write out explicitly
what "it" is referring to.\\
4. Do not guess anything and if you are not sure then do not change it.\\
5. If nothing happens between two actions of [Inferred\_agent] then treat
them as one action.\\

The output format should be: ["aaa.", "bbb.", ...], ["aaa.", "bbb.", ...],
and should be the same length as the list [bare\_actions]. Output only this
list.\\
\mbox{[Normalized action list]}
\end{PromptBox}

\begin{PromptBox}{Single-action wording repair}
Here is an action/utterance of [Inferred\_agent]: [Sentence]\\
If it is a full sentence, output the same thing as the input. Do not modify
the original punctuation.\\
If it is not in full sentence, rephrase it so it has [Inferred\_agent]. Make
sure that it is grammatically correct, without losing any information. Do
not change the original wording.\\
Just output a sentence without explanation.\\
Output:
\end{PromptBox}

\begin{PromptBox}{Sentence-level nested-perspective test}
Determine if [Agent] can observe the sentence in the story.\\
{[Agent]} can only observe events where [Agent] is located in. [Agent] can
observe everything (including objects inside containers) in their
location.\\
After [Agent] exits a room, [Agent] cannot observe events in the room.\\
If the sentence is about someone's inner thoughts, [Agent] cannot observe
it.\\

{[Story]}\\

Determine if [Agent] can observe: [Sentence]\\
A) [Agent] can observe the sentence.\\
B) [Agent] cannot observe the sentence.\\
Answer:
\end{PromptBox}

The action templates are executed sequentially: verbatim extraction, batch
wording normalization, and a sentence-level repair pass. Nested
reconstruction applies the A/B visibility test independently to each
sentence. Participation changes are retained as source-tagged events so that
the ordered observer chain remains available to higher-order reconstruction.

\LinkTarget{jump-schema}
\subsubsection{Shared Schema and Task-Aware Instantiation}

Prompt outputs and benchmark-provided event fields are normalized into the
shared schema in Table~\ref{tab:parser_rules}. This stage resolves surface
forms, aliases, and native field layouts while preserving the agents, entities,
relations, temporal order, and evidence sources supplied by the input. Because
the benchmarks expose different mental-state queries and event structures, the
instantiation is task-aware: it selects the inferred variable and the
structured relations required by that question family. This follows the
benchmark-interface design of AutoToM; it does not
introduce answer labels or
candidate-dependent calibration. The same query frame is applied to every
candidate in an example, preventing candidate-dependent reinterpretation of
the question.

\begin{table*}[t]
\centering
\small
\setlength{\tabcolsep}{5pt}
\renewcommand{\arraystretch}{1.10}
\begin{tabularx}{0.98\textwidth}{@{}p{0.16\textwidth}p{0.38\textwidth}X@{}}
\toprule
\textbf{Schema field} & \textbf{Input cue and normalization} &
\textbf{Normalized output} \\
\midrule
Mental-state type $\tau$ &
Question predicates and query form, including \emph{think}, \emph{believe},
\emph{trying to get}, social-effect relations, and nested attribution clauses &
\textsc{State}, \textsc{Belief}, \textsc{Goal}, \textsc{SocialGoal}, or
\textsc{BeliefOfGoal} \\
Perspective holder $i$ &
Named subject of the queried mental state; for nested questions, the final
holder in the parsed attribution chain &
Canonical agent name \\
Target agent $j$ &
Listener, affected agent, or the agent whose goal is attributed by another
agent &
Canonical agent name or \textsc{None} \\
Entity $x$ &
Object or location shared by the question and candidates; aliases are
normalized by removing articles and spacing variants &
Canonical object or location string \\
Question relation $\rho$ &
Lexical operator and relation in the question &
\textsc{Most}, \textsc{Least}, positive containment, negative containment,
or nested attribution \\
Candidate value $\chi_k$ &
Candidate-specific object, location, belief value, or goal object &
Typed candidate value \\
Candidate effect $\epsilon_k$ &
Social-effect markers and the relation between an action or utterance and
another agent's goal &
\textsc{Help}, \textsc{Hinder}, \textsc{Indifferent}, or \textsc{None} \\
\bottomrule
\end{tabularx}
\caption{Shared schema fields and input-normalization interface.}
\label{tab:parser_rules}
\end{table*}

\paragraph{Input adapters.}
The adapters retain the information already exposed by each benchmark, such
as typed candidate values, structured state or action fields, and ordered
attribution chains. Native expressions are converted into common observation,
action, and utterance records while preserving agent, entity, time, and source
tags. The agent-specific perspective store, Bayesian inference backbone, and
multiplicative factor interface are shared across benchmarks. The concrete
relation extractor and route-activation condition are selected by the
normalized task type, so task-aware instantiation should not be read as a
separate fitted factor set for each benchmark.

\LinkTarget{jump-store}
\subsubsection{Agent-Specific Perspective Storage}

For each event and agent, the implementation evaluates visibility,
audibility, and self-action cues. An admitted event is stored together with
its source. Communication remains an utterance event and is not converted
directly into a physical world-state fact.

\begin{algorithm}[t]
\caption{Agent-Specific Perspective Storage.}
\label{alg:perspective_store}
\begin{algorithmic}[1]
\REQUIRE Events $E_{1:T}$ and agent set $\mathcal{A}$
\ENSURE Final perspective stores $\{\store_i^T:i\in\mathcal{A}\}$
\FOR{agent $i\in\mathcal{A}$}
  \STATE Initialize
  $\store_i^0=\{O_i=\emptyset,A_i=\emptyset,U_i=\emptyset,M_i=\emptyset\}$
\ENDFOR
\FOR{$t=1,\ldots,T$}
  \FOR{agent $i\in\mathcal{A}$}
    \STATE $\store_i^t\leftarrow\store_i^{t-1}$
    \STATE $g_i^t\leftarrow
    \mathbf{1}[v_i(e_t)\vee h_i(e_t)\vee a_i(e_t)]$
    \IF{$g_i^t=1$}
      \STATE Append the event to the corresponding observation, action, or
      utterance slot
      \STATE Preserve actor, speaker, listener, timestamp, and visibility
      source tags
      \STATE $M_i^t\leftarrow
      \mathrm{UpdateMemory}(M_i^{t-1},e_t,\mathrm{source}(e_t))$
    \ENDIF
  \ENDFOR
\ENDFOR
\RETURN $\{\store_i^T\}$
\end{algorithmic}
\end{algorithm}

The serialized store records own actions, observed actions, own utterances,
heard utterances, observations, communication events, and source-aware
memory. Derived belief estimates can be cached beside this evidence, but they
retain an inference label and are not reintroduced as direct observations.
For physical belief and state scoring, the memory renderer can omit
communicative content; for Social Goal and Belief of Goal scoring, it renders
the same communication with explicit speaker and listener tags.

\LinkTarget{jump-input}
\subsubsection{Input Construction}
\label{sec:input_construction}

\paragraph{Fused event traces.}
\pip{} operates on a fused event trace rather than raw frames. Textual
episodes are aligned with any benchmark-provided visual actions, state
sequences, or action lists and converted into the observation, action, and
utterance records described in Section~\ref{sec:parsing_prompts}. This
normalization preserves the supplied content and temporal order; it does not
alter the candidate set or introduce task-dependent evidence weights.

\paragraph{Candidate order.}
Candidates are shuffled with a fixed random seed before inference. Ground-truth
indices are recomputed after shuffling. The same candidate order is used by the full
model and its ablations for a given backbone and sample.

\paragraph{No answer supervision.}
Ground-truth answers are used only for evaluation. Perspective stores,
schemas, route activation, and factors are constructed from the episode,
question, and candidate text. No sample-specific factor fitting or validation
on the answer label is performed.

\LinkTarget{jump-routes}
\subsection{Route Activation, Aggregation, and Factor Ranges}
\label{sec:factor_ranges}

We use the four evidence routes defined in the main paper---Perceptual,
Behavioral, Communicative, and Recursive---without introducing additional
route types or scoring equations. The factor extraction, multiplication, and
likelihood calibration follow Equations~(8)--(10) of the main paper. This section
therefore reports only the operating ranges omitted from the main text.

\begin{ImplementationBox}{Route-specific factor operating ranges}
\small

\textbf{Perceptual evidence:} Factors for accessible physical states,
localization cues, and initial observations operate approximately within
$[0.45,2.0]$.

\textbf{Behavioral evidence:} Factors for search behavior, goal-directed action,
and trajectory consistency operate approximately within $[0.34,1.75]$.

\textbf{Communicative evidence:} Communication-effect factors use the discrete
set $\{0.01,0.08,1.00,1.08\}$, while the combined
utterance--belief relation lies approximately within $[0.15,1.30]$.

\textbf{Recursive evidence:} Factors for nested belief and goal attribution
operate approximately within $[0.15,1.35]$.

\end{ImplementationBox}

These factor families are shared across tasks, with all operating ranges fixed
at the run level rather than fitted to individual instances. The parsed claim
schema selects the applicable subset of the four routes and their corresponding
ranges. A factor of $1$ is neutral; values below $1$ downweight incompatible
evidence, whereas values above $1$ provide bounded support. Claim-dependent
route activation and multi-factor composition are described in
Section~\ref{sec:task_instantiations}.

\LinkTarget{jump-decision}
\subsection{Task Instantiations and Evidence Handling}
\label{sec:task_instantiations}

The four route names denote complementary evidence types rather than
mutually exclusive task labels. Each structured claim is evaluated by
combining the applicable route-specific factors. Table~\ref{tab:task_routes} summarizes these multi-factor compositions; an optional route
contributes the neutral factor when its corresponding evidence source is
absent or inapplicable.

\begin{table*}[t]
\centering
\footnotesize
\setlength{\tabcolsep}{2pt}
\renewcommand{\arraystretch}{1.12}

\begin{tabularx}{0.95\textwidth}{
@{}
>{\raggedright\arraybackslash}p{0.21\textwidth}
>{\raggedright\arraybackslash}p{0.27\textwidth}
>{\raggedright\arraybackslash}X
@{}
}
\toprule
\textbf{Claim structure}
&
\textbf{Route-factor composition}
&
\textbf{Evidential roles}
\\
\midrule

Physical state or lower-order belief
&
Perceptual $\times$ Behavioral $\times$ Communicative
&
Accessible states and state changes are combined with action consistency.
A source-tagged utterance contributes when it is relevant to the queried
belief.
\\

\addlinespace[4pt]

Goal from observed behavior
&
Behavioral $\times$ Perceptual $\times$ Communicative
&
Search, movement, and interaction evidence is grounded in the agent-accessible
object or location state. A relevant request or utterance may provide
additional goal evidence.
\\

\addlinespace[4pt]

Social Goal
&
Communicative $\times$ Perceptual $\times$ Behavioral
&
The source-tagged utterance is evaluated jointly with the speaker's accessible
belief and its intended effect on the listener's action.
\\

\addlinespace[4pt]

Belief of Goal or higher-order belief
&
Recursive $\times$ Perceptual $\times$ Behavioral $\times$
Communicative
&
The attribution chain is combined with state and action evidence, together
with utterance evidence when it supports the inner claim.
\\

\bottomrule
\end{tabularx}

\caption{Multi-route factor composition by claim structure.
We treat orders~0--1 as lower-order and orders~2 and above as higher-order.}
\label{tab:task_routes}
\end{table*}

Thus, route activation is not a single-route choice: each claim combines
multiple complementary factors, while unavailable optional evidence leaves
the product unchanged through the neutral factor.

\paragraph{Mutually exclusive branches.}
Within one route and one local term, alternative interpretations of the same
evidence record are mutually exclusive. Only the branch compatible with
$R_q$ and $Z_k$ contributes a factor; its alternative branches are not
multiplied again. This prevents one event from being counted simultaneously as
both support and counterevidence.

\paragraph{Independent evidence and conflicts.}
Factors from distinct admitted evidence records or complementary routes are
multiplied as in Equation~(9) of the main paper. Supporting evidence contributes a
factor above $1$, while explicit counterevidence contributes a factor below
$1$. If independent sources disagree, neither source is discarded: their
support and downweighting are combined in the same local term before the
likelihood cap is applied.

\paragraph{Missing and uncertain evidence.}
An inactive route, an unavailable evidence field, or a relation that cannot be
reliably determined contributes the neutral factor $1$. Missing evidence is
not treated as counterevidence by default. It produces a downweighting factor
only when the instantiated claim makes the absent observation or action itself
diagnostic. Ambiguous evidence likewise remains neutral rather than being
assigned a guessed direction.

\paragraph{Hypothesis marginalization.}
Calibration changes only the route-relevant local terms. The underlying model
still sums over alternative latent assignments and then normalizes across
candidates, as defined in Equation~(11) of the main paper. The \textsc{Most} and
\textsc{Least} decoding rules are unchanged.

\LinkTarget{jump-experiments}
\section{Additional Results and Experimental Details}
\label{sec:experiment_details}
\LinkTarget{jump-baselines}
\subsection{Baseline Implementation Details}
\label{sec:baseline_details}
\paragraph{Reporting protocol.}
Unless otherwise stated, baseline results are taken from the corresponding papers under their originally reported backbones, datasets, and evaluation settings.

\paragraph{General-purpose LLMs.}
Gemini-3.5-Flash,
GPT-4o, GPT-5.4,
Qwen3-VL-8B, and
InternVL3.5-8B serve
both as direct question-answering baselines and as backbones for \pip{}. In our
matched backbone comparisons, temperature, answer format, candidate order,
and multimodal input construction are held fixed, so the differences primarily
reflect the reasoning framework applied on top of the same backbone.

\paragraph{ToM-specific reasoning scaffolds.}
SimToM filters the episode according to the target
agent's accessible observations before answering from the resulting
perspective-conditioned context. ThoughtTracing
constructs an explicit,
hypothesis-driven reasoning trace before answer selection. The reported
results are taken directly from the original papers under their published
backbones and evaluation settings. ThoughtTracing is included only for
text-based benchmarks and tasks supported by its original interface.

\paragraph{State-tracking methods.}
DWM and PDDL-Mind
explicitly reconstruct world and belief states before answer selection. Their
original implementations operate on textual event descriptions and do not
provide a native visual front end. We therefore retain their reported
text-based results on compatible benchmarks rather than adding a new visual
encoder or modifying their original state representations.

\paragraph{Model-based mental-state inference.}
BIP-ALM and LIMP combine
Bayesian inverse planning with language models. BIP-ALM manually specifies
symbolic observable and latent variables under a POMDP formulation, whereas
LIMP represents variables in natural language and is designed primarily for
two-level reasoning. Both methods assume object-location goals and beliefs in
household-style environments, limiting their direct applicability to more
open-ended or higher-order tasks. AutoToM
automatically discovers task-relevant variables and constructs the
corresponding Bayesian model. Its published results are retained where
available. Since the original AutoToM paper does not report results with
Qwen3-VL-8B or InternVL3.5-8B, we reproduce AutoToM with these two backbones
using its released implementation. In these reproduced comparisons, AutoToM
and \pip{} use the same fused video--text event interface and matched backbone settings,
focusing the comparison on perspective construction and evidence
calibration.

\paragraph{Direct-prompt protocols.}
The following unified prompt is used to evaluate the open-source models:
comments mark dataset-specific fields and instructions.

\begin{BaselinePromptBox}{Open-source direct prompt}
Background Story:\\
\{story\}\\[2pt]
\textcolor{PIPGreen}{\textit{\# MMToM-QA only}}\\
Current States:\\
\{states\}\\[2pt]
Character Actions:\\
\{actions\}\\[2pt]
Question:\\
\{question\}\\[2pt]
Answer Choices:\\
\{formatted\_choices\}\\[2pt]
\textcolor{PIPGreen}{\textit{\# MuMA-ToM and MMToM-QA}}\\
--- Instruction ---\\
Analyze the character's hidden belief, intention, or knowledge state based on
the constraints. Write a very concise reasoning within 2-3 sentences. Do not
repeat the background text.\\
Now, verify each option objectively against your reasoning above. Output ONLY
the one correct chosen option text exactly, without any extra words, list
symbols, or explanation.\\[3pt]
\textcolor{PIPGreen}{\textit{\# Hi-ToM only (standalone evaluation)}}\\
--- Instruction ---\\
Analyze the scenario and reason step-by-step based on the constraints. Write a
very concise reasoning within 2-3 sentences. Do not repeat the background
text.\\[2pt]
Your Analysis:\\
\{cot\_reasoning\}\\[2pt]
--- New Instruction ---\\
Now, verify each option objectively against your reasoning above. Output ONLY
the one correct chosen option text exactly, without any extra words, list
symbols, or explanation.
\end{BaselinePromptBox}

The following prompt is used to evaluate the closed-source model.

\begin{BaselinePromptBox}{Closed-source direct prompt}
[System]\\
Answer the question using only the given story.\\[3pt]
[User]\\
Story:\\
\{story\}\\[2pt]
Question:\\
\{question\}\\[2pt]
Options:\\
\{formatted\_choices\}\\[2pt]
End your reasoning with exactly one line: Final answer: <one of
\{allowed\}>
\end{BaselinePromptBox}

\LinkTarget{jump-benchmarks}
\subsection{Benchmark Details}
\label{sec:benchmark_details}

The main paper summarizes the three benchmarks. Additional evaluation details
are given below.

\paragraph{MuMA-ToM.}
Each question has three answer candidates. Social Goal questions distinguish
helping, hindering, and independent behavior; Belief of Goal questions require
one agent to infer another's object-directed goal.

\paragraph{MMToM-QA.}
We use the multimodal setting. The Belief category contains 100 questions per
subtype: true belief, short-term (1.1); false belief, short-term (1.2); and
belief tracking, long-term (1.3). The Goal category contains 75 questions per
subtype: goal given true belief (2.1), false belief (2.2), updated belief (2.3),
and future actions (2.4).

\paragraph{Hi-ToM.}
Each order of the evaluated length-1 subset contains 20 instances with
communication and 20 without. Each original instance provides 15 location
candidates. Following AutoToM, we retain candidates whose location tokens occur
in the episode and remap the ground-truth index to the filtered list.

\LinkTarget{jump-additional-experiments}
\subsection{Additional Experiments}
\label{sec:additional_experiments}

\LinkTarget{jump-detailed-experimental-results}
\subsubsection{Detailed Experimental Results}

Table~\ref{tab:detailed_results_combined} aligns the subtype- and order-level
results from two matched-backbone studies. The Gemini-3.5-Flash columns compare
direct prompting with \pip{} on MMToM-QA and Hi-ToM, while the Qwen3-VL-8B
columns compare the full framework with \textit{w/o Perspective Storage} on all
three benchmarks. A dash denotes a comparison that was not run.

\begin{table*}[t]
\centering
\small
\setlength{\tabcolsep}{5pt}
\renewcommand{\arraystretch}{1.06}

\resizebox{\textwidth}{!}{%
\begin{tabular}{@{}llcccc@{}}
\toprule
\textbf{Benchmark}
& \textbf{Subset}
& \multicolumn{2}{c}{\textbf{Gemini-3.5-Flash}}
& \multicolumn{2}{c}{\textbf{Qwen3-VL-8B}} \\
\cmidrule(lr){3-4}
\cmidrule(lr){5-6}
&
& \multicolumn{1}{c}{\textbf{Direct}}
& \multicolumn{1}{c}{\textbf{\pip{}}}
& \multicolumn{1}{c}{\textbf{Full}}
& \multicolumn{1}{c}{\textbf{w/o Perspective Storage}} \\
\midrule

\multirow{4}{*}{MuMA-ToM}
& Belief
& -- & -- & 91.33 & 84.33 \\
& Social Goal
& -- & -- & 82.00 & 80.00 \\
& Belief of Goal
& -- & -- & 82.00 & 84.67 \\
& \textbf{Overall}
& -- & -- & \textbf{85.11} & \textbf{83.00} \\

\midrule

\multirow{10}{*}{MMToM-QA}
& True belief, short-term (1.1)
& 94.00 & 100.00 & 99.00 & 95.00 \\
& False belief, short-term (1.2)
& 100.00 & 100.00 & 94.00 & 67.00 \\
& Belief tracking, long-term (1.3)
& 94.00 & 100.00 & 78.00 & 77.00 \\
& \textbf{Belief average}
& \textbf{96.00}
& \textbf{100.00}
& \textbf{90.33}
& \textbf{79.67} \\

\cmidrule(lr){2-6}

& Goal given true belief (2.1)
& 66.67 & 96.00 & 96.00 & 94.67 \\
& Goal given false belief (2.2)
& 90.67 & 88.00 & 72.00 & 69.33 \\
& Goal given updated belief (2.3)
& 2.67 & 98.67 & 85.33 & 73.33 \\
& Goal given future actions (2.4)
& 68.00 & 81.33 & 58.67 & 56.00 \\
& \textbf{Goal average}
& \textbf{57.00}
& \textbf{91.00}
& \textbf{78.00}
& \textbf{73.33} \\
& \textbf{Overall}
& \textbf{76.50}
& \textbf{95.50}
& \textbf{84.17}
& \textbf{76.50} \\

\midrule

\multirow{6}{*}{Hi-ToM}
& Order~0
& 100.00 & 100.00 & 100.00 & 100.00 \\
& Order~1
& 95.00 & 100.00 & 70.00 & 72.50 \\
& Order~2
& 92.50 & 97.50 & 95.00 & 80.00 \\
& Order~3
& 85.00 & 97.50 & 92.50 & 87.50 \\
& Order~4
& 87.50 & 92.50 & 77.50 & 70.00 \\
& \textbf{Overall}
& \textbf{92.00}
& \textbf{97.50}
& \textbf{87.00}
& \textbf{82.00} \\

\bottomrule
\end{tabular}%
}

\caption{
Detailed accuracy (\%) for the matched Gemini-3.5-Flash and
Qwen3-VL-8B comparisons.
}
\label{tab:detailed_results_combined}
\end{table*}

\pip{} yields its largest Gemini gain on \emph{Goal given updated belief (2.3)}.
The \textit{w/o Perspective Storage} ablation most clearly degrades
Qwen3-VL-8B on \emph{False belief, short-term (1.2)} and
\emph{Goal given updated belief (2.3)}, as well as on
higher-order Hi-ToM subsets.

\LinkTarget{jump-token-consumption}
\subsubsection{Token Consumption Analysis}
\label{sec:token_consumption}
Table~2 of the main paper reports dataset-level token costs. Here we provide
subtask-level averages in thousands of total prompt and completion tokens per
question (K), computed over completed inference instances.

\Needspace{0.45\textheight}
\paragraph{Subtask-level comparison.}
Table~\ref{tab:token_subtasks} reports the available subtask averages by
configuration and backbone. Under \textit{w/o Perspective Storage}, Qwen3-VL-8B is most expensive
on MuMA-ToM Social Goal (49.47K), while Hi-ToM order~1 is the most
token-intensive order for both backends. Under
\textit{w/o Claim-Aware Calibration}, MuMA-ToM Social Goal
is again the largest average, although it is computed from 90 available
questions. For AutoToM, Belief of Goal is the most token-intensive MuMA-ToM
subtask for both open-source backends. We report observed averages without
extrapolating missing questions.

\begin{table*}[t]
\centering
\small
\setlength{\tabcolsep}{4.5pt}
\renewcommand{\arraystretch}{1.15}

\resizebox{\textwidth}{!}{%
\begin{tabular}{@{}lllccccc ccccc@{}}
\toprule
\multirow{2}{*}{\textbf{Method}}
& \multirow{2}{*}{\textbf{Setting}}
& \multirow{2}{*}{\textbf{Backend}}
& \multicolumn{2}{c}{\textbf{MMToM-QA}}
& \multicolumn{3}{c}{\textbf{MuMA-ToM}}
& \multicolumn{5}{c}{\textbf{Hi-ToM}} \\
\cmidrule(lr){4-5}
\cmidrule(lr){6-8}
\cmidrule(lr){9-13}
&
&
& \textbf{Belief}
& \textbf{Goal}
& \textbf{Belief}
& \textbf{Social Goal}
& \textbf{Belief of Goal}
& \textbf{O0}
& \textbf{O1}
& \textbf{O2}
& \textbf{O3}
& \textbf{O4} \\
\midrule

\multirow{3}{*}{\pip{}}
& \multirow{2}{*}{w/o Perspective Storage}
& GPT-4o
& 3.89
& 5.94
& 12.97
& 18.20
& 7.89
& 15.47
& 52.45
& 23.44
& 13.56
& 8.92 \\

&
& Qwen3-VL-8B
& 3.22
& 4.31
& 26.59
& 49.47
& 29.53
& 15.19
& 70.37
& 30.43
& 17.04
& 10.12 \\

\cmidrule(lr){2-13}

& w/o Claim-Aware Calibration
& GPT-4o
& 4.04
& 7.84
& 13.15
& 21.60
& 7.87
& --
& --
& --
& --
& -- \\

\midrule

\multirow{2}{*}{AutoToM}
& \multirow{2}{*}{Full}
& Qwen3-VL-8B
& 4.23
& 8.01
& 12.08
& 14.79
& 20.31
& --
& --
& --
& --
& -- \\

&
& InternVL3.5-8B
& 3.36
& 8.19
& 4.02
& 5.37
& 14.20
& --
& --
& --
& --
& -- \\

\bottomrule
\end{tabular}%
}

\caption{
Subtask-level average total tokens per question, measured in thousands (K).
O0--O4 denote reasoning Orders 0--4 on Hi-ToM.
}
\label{tab:token_subtasks}
\end{table*}

\LinkTarget{jump-random-seed-robustness}
\subsubsection{Robustness to Random Seeds}

To assess the sensitivity of our results to random seeds, we evaluated
\pip{} with InternVL3.5-8B using three different seeds: 42, 1234, and
2024.

\begin{table*}[t]
\centering
\small
\setlength{\tabcolsep}{3.5pt}
\renewcommand{\arraystretch}{1.04}
\resizebox{\textwidth}{!}{%
\begin{tabular}{@{}c cccc ccc@{}}
\toprule
\multirow{2}{*}{\textbf{Seed}}
& \multicolumn{4}{c}{\textbf{MuMA-ToM}}
& \multicolumn{3}{c}{\textbf{MMToM-QA}} \\
\cmidrule(lr){2-5} \cmidrule(lr){6-8}
& Belief & \shortstack{Social\\Goal} &
\shortstack{Belief of\\Goal} & Avg.
& Belief & Goal & Avg. \\
\midrule
42   & 93.67 & 83.67 & 74.33 & 83.89
     & 93.33 & 81.33 & 87.33 \\
1234 & 93.33 & 83.67 & 75.33 & 84.11
     & 91.67 & 81.67 & 86.67 \\
2024 & 93.67 & 84.00 & 75.33 & 84.33
     & 91.67 & 81.67 & 86.67 \\
\midrule
Mean $\pm$ SE
     & $93.56 \pm 0.11$
     & $83.78 \pm 0.11$
     & $75.00 \pm 0.33$
     & $\mathbf{84.11 \pm 0.13}$
     & $92.22 \pm 0.56$
     & $81.56 \pm 0.11$
     & $\mathbf{86.89 \pm 0.22}$ \\
\bottomrule
\end{tabular}%
}
\caption{Accuracy (\%) across random seeds for \pip{} with InternVL3.5-8B;
the last row reports mean $\pm$ standard error.}
\label{tab:random_seed_robustness}
\end{table*}

As shown in Table~\ref{tab:random_seed_robustness}, \pip{} achieved a mean accuracy of 84.11\% with a standard error of 0.13\% on MuMA-ToM, closely matching the 83.89\% obtained with seed~42 and reported in the main results. Similarly, its mean accuracy on MMToM-QA was 86.89\% with
a standard error of 0.22\%, consistent with the reported seed~42 result of 87.33\%. Across the individual subtasks, the standard errors ranged
from only 0.11\% to 0.56\%. Moreover, the overall accuracy varied by at most 0.44 percentage points on MuMA-ToM and 0.66 percentage points on MMToM-QA across the three runs. These results demonstrate that the
performance of \pip{} is stable under the evaluated random-seed settings and that our main conclusions remain consistent across runs.

\LinkTarget{jump-audit}
\section{Manual Audit of Perspective-Boundary Errors}
\label{sec:manual_audit}

\LinkTarget{jump-audit-scope}
\subsection{Audit Scope and Label Hierarchy}

We compare paired GPT-4o predictions from the \textit{Full} and
\textit{w/o Perspective Storage} configurations, generated with identical
inputs, candidate order, seed, schema, calibration, and Bayesian inference backbone. A
case enters the Full-only pool when the \textit{Full} configuration is correct and
the ablation is incorrect. The paired runs yield 151 such cases.

We analyze the 30 MuMA-ToM Social Goal cases separately because their candidates
express intentions or social outcomes rather than directly comparable physical
states. The remaining 121 cases constitute the strictly comparable audit pool.

Each retained case receives a broad binary label indicating whether the
ablated prediction relies on inaccessible/global evidence. When the
candidate semantics permit direct comparison with the objective state, we
additionally record literal and direction-consistent objective-state matches.
The mechanism decomposition is reported in Table~6 of the main paper.

\LinkTarget{jump-audit-procedure}
\subsection{Audit Procedure}

\begin{algorithm}[t]
\caption{Layered perspective-boundary audit.}
\label{alg:manual_audit}
\begin{algorithmic}[1]
\REQUIRE The 121 strictly comparable Full-only prediction pairs
\ENSURE Broad inaccessible/global evidence label and mechanism-specific diagnostic
\FOR{each paired sample}
    \STATE Align the sample ID, question, choices, and output artifacts
    \STATE Reconstruct the objective and agent-accessible event sequences
    \STATE Identify the evidence or relation supporting the ablated answer
    \STATE Assign a binary label for inaccessible/global evidence
    \IF{the candidates permit objective-state comparison}
        \STATE Record whether the selected candidate literally matches the
        objective state
        \STATE Apply the \textsc{Most}/\textsc{Least} operator and record
        direction consistency
    \ENDIF
    \STATE Record the corresponding relation, grounding, polarity, path, or
    parsing error type
\ENDFOR
\RETURN Case-level labels and aggregate counts
\end{algorithmic}
\end{algorithm}

For \textsc{Least} questions, a literal objective-state match is counted as
direct objective-state attraction only when the selected option is also
consistent with the question direction. This rule prevents polarity reversals from inflating the
perspective-leakage estimate.